\documentclass[Afour,sageh,times]{sagej}

\usepackage{moreverb,url}
\usepackage[colorlinks,bookmarksopen,bookmarksnumbered,citecolor=red,urlcolor=red, draft]{hyperref}

\usepackage{subfigure}
\usepackage{graphicx}
\usepackage{algorithmic}
\usepackage{algorithm}
\usepackage{amsmath}
\usepackage{verbatim}
\usepackage{amsthm}
\usepackage{amsfonts}
\usepackage{amssymb}
\usepackage{enumitem}
\usepackage{placeins}
\usepackage[normalem]{ulem}

\usepackage{array,arydshln}

\usepackage{titlesec}
\usepackage{mathtools}
\def\multiset#1#2{\ensuremath{\left(\kern-.3em\left(\genfrac{}{}{0pt}{}{#1}{#2}\right)\kern-.3em\right)}}

\newtheoremstyle{dotless}{}{}{\itshape}{}{\bfseries}{}{ }{}
\theoremstyle{dotless}
\newtheorem{theorem}{Theorem}
\newtheorem{remark}{Remark} 

\newcommand\redsout{\bgroup\markoverwith{\textcolor{red}{\rule[0.5ex]{2pt}{0.4pt}}}\ULon}

\newcommand\BibTeX{{\rmfamily B\kern-.05em \textsc{i\kern-.025em b}\kern-.08em
		T\kern-.1667em\lower.7ex\hbox{E}\kern-.125emX}}

\DeclareMathOperator*{\argmin}{argmin}
\def\volumeyear{2017}
\begin{document}
	
	\runninghead{Kapoutsis, et al.}
	
	\title{A distributed, plug-n-play algorithm for multi-robot applications with a priori non-computable objective functions}
	
	\author{Athanasios Ch. Kapoutsis\affilnum{1,2}, Savvas A. Chatzichristofis\affilnum{3} and Elias B. Kosmatopoulos\affilnum{1,2}}
	\affiliation{\affilnum{1}Department of Electrical and Computer Engineering,	Democritus University of Thrace, Xanthi, Greece \\
		\affilnum{2}Information Technologies Institute, The Centre for Research \& Technology, Hellas, Thessaloniki, Greece \\
		\affilnum{3}Neapolis University, Department of Computer Science, Paphos, Cyprus}
	
	\corrauth{Athanasios Ch. Kapoutsis, Department of Electrical and Computer Engineering,	Democritus University of Thrace, Xanthi, 67100, Greece.}
	
	\email{akapouts@ee.duth.gr}
	
	\begin{laysummary}
		\begin{center}
		 \href{https://github.com/athakapo/A-distributed-plug-n-play-algorithm-for-multi-robot-applications}{https://github.com/athakapo/A-distributed-plug-n-play-algorithm-for-multi-robot-applications}
		 \end{center}
	\end{laysummary}
	
	\begin{abstract}
		This paper presents a distributed algorithm applicable to a wide range of practical multi-robot applications. In such multi-robot applications, the user-defined objectives of the mission can be cast as a general optimization problem, without explicit guidelines of the subtasks per different robot. Owing to the unknown environment, unknown robot dynamics, sensor nonlinearities, etc., the analytic form of the optimization cost function is not available a priori. Therefore, standard gradient-descent-like algorithms are not applicable to these problems. To tackle this, we introduce a new algorithm that carefully designs each robot's subcost function, the optimization of which can accomplish the overall team objective. Upon this transformation, we propose a distributed methodology based on the cognitive-based adaptive optimization (CAO) algorithm, that is able to approximate the evolution of each robot's cost function and to adequately optimize its decision variables (robot actions). The latter can be achieved by online learning only the problem-specific characteristics that affect the accomplishment of mission objectives. The overall, low-complexity algorithm can straightforwardly incorporate any kind of operational constraint, is fault tolerant, and can appropriately tackle time-varying cost functions. A cornerstone of this approach is that it shares the same convergence characteristics as those of block coordinate descent algorithms. The proposed algorithm is evaluated in three heterogeneous simulation set-ups under multiple scenarios, against both general-purpose and problem-specific algorithms.
	\end{abstract}

	
	\maketitle
	
	\section{Introduction}
	\label{sec:intro}
	
	The new era of artificial intelligence and robotics has an ever-increasing interest in multi-robot systems. \color{black} The causality of this trend is outlined in the following three points. First, the recent \textit{advances \color{black} in hardware and communications} allow the cooperative deployment of many affordable robots. Second, the use of multiple robots introduces \textit{redundancy}, which can be translated into mission speed-up and/or fault-tolerant characteristics (e.g., in cases when one or more robots faces a malfunction). Third, the utilization of multi-robot teams may tackle \textit{problems that cannot be solved with a single robot} (e.g., continuous monitoring/guarding a large area).  Robot missions in which the multi-robot configuration can be more appealing include surveillance in hostile environments (e.g., areas contaminated with biological, chemical, or even nuclear wastes), law enforcement missions (e.g., border patrol), agriculture activities (e.g., soil sampling), and cleaning missions (e.g., cleaning up an oil spill).    
	
	\subsection{Related work}
	\label{subsec:related}
	
	Unfortunately, many of the multi robot tasks have been proven to be extremely difficult. For example, the online generation of robot trajectories so as to maximize SLAM accuracy and efficiency is NP-hard \cite[]{singh2009efficient, kollar2008efficient}. Moreover, the offline design of multi-robot trajectories in order to cover a known area of interest in minimum time/energy has been proven NP-complete \cite[]{zheng2005multi}, etc. 
	
	To alleviate the above problem, many multi-robot approaches attempt to solve  a simplified version of the original problem.  In such a way, it is possible to construct a computationally feasible solution, utilizing \textit{optimal control or dynamic programming} techniques, at the expense, of course, of sacrificing global optimality. For instance, to render the decision-making scheme computationally feasible, many methodologies  \cite[]{seyboth2015robust, le2009trajectory, de2008dynamic} assumed relaxed or linearized versions of the multi-robot problem. A usual assumption is that the robots operate in a discrete space where their actions and measurements can also take values from a finite discrete set of values \cite[]{matignon2012coordinated, spaan2005perseus}. The exploitation of the above assumption can lead to remarkable results in the context of multi-robot tasks, presenting many real-life applications (e.g., \cite{capitan2013decentralized}). Unfortunately, these strategies cannot be fully informed by the (usually occurring) continuous field measurements, whereas they can be computationally intractable for large state systems, e.g., a single mobile robot operating in the real world often has millions of possible states \cite[]{roy1999coastal}. Other multi-robot approaches that fall into this class adapt the assumption of perfect or sufficient knowledge of the dynamics of the overall multi-robot system, i.e., the dynamics of each and every robot along with their interactions with the other robots and the external environment \cite[]{wang2016multi,  zhou2011multirobot}. In such cases, the multi-robot problem can be seen to be equivalent to a standard optimization problem, where the robots' decision values are generated according to, e.g., a  gradient-descent or gradient-descent-like algorithm \cite[]{nesterov2007gradient}. However, the requirement for perfect or sufficient knowledge of the overall dynamics renders the overall control design practically infeasible in many multi-robot applications, as they typically involve a large number of controllable variables with highly complex and uncertain dynamics \cite[]{morgan2016swarm, chen2015occlusion, gomes2013evolution}. 
	
	Another well-investigated class of multi-robot approaches is the \textit{optimal one-step-ahead} methodologies. In this family of approaches, the next robots' decision variables are chosen greedily, so as to optimize an appropriately defined cost function that is related to the problem in hand. For instance, in the domain of multi-robot exploration, a common practice  is to choose the next robots' positions that maximize the expected information gain \cite[]{rooker2007multi, burgard2005coordinated, stachniss2003exploring} or minimize the trace of the extended Kalman filter (EKF) error covariance matrix \cite[]{cui2016mutual, bourgault2002information}. Although, many of these approaches have been successfully evaluated in real-life multi-robot platforms, the majority of them suffer from the following drawbacks. First and foremost, the nonlinearities may give rise to undesirable divergence (such as in cases where the noise does not follow the additive white Gaussian noise (AWGN) model). For example, it is usually considered that a robot can accurately estimate the position of an object or a point in the environment (landmark/cell) as soon as it perceives it. In most of the existing \textit{optimal-one-step-ahead} approaches, this assumption allows in each timestamp the a priori calculation of the cost function, as well as the robots' decision variables that greedily optimize such a cost function. Moreover, such an assumption is crucial for overcoming deadlocks (local minima), which are frequently encountered when greedy approaches are employed \cite[]{palacios2016distributed,rathnam2013distributed}. Finally, the selection of an adequate cost function that provides an efficient solution to the multi-robot problem is not always trivial. 
	
	On the other side of the spectrum are the \textit{simulation-based} multi-robot methodologies \cite[]{kapoutsis2015noptilus, kollar2008trajectory, kohl2004policy}. The idea behind these approaches is as follows. First a parameterized decision-making mechanism is devised for generating the robot decisions online, with different choices for its parameters, leading to different decision-making mechanisms. Then, realistic simulations or similar tools are used in order to optimize the parameters of the decision-making mechanism. Thus, conceptually, many of the optimization computations that otherwise would take place on the real devices are ``moved'' offline. The drawbacks of such approaches are as follows: first, the simulations need to cover a wide range of different realistic scenarios (and, thus, they may become ``expensive'') and second, because the dimensionality of the optimization problem is quite high, a large number of parameters is needed in order to come up with an efficient decision-making mechanism. 
	
	We close this subsection by mentioning that for most of the centralized approaches, in all three classes, it is not clear how they can be extended to have a distributed nature. Furthermore, the majority of the distributed multi-robot algorithms [e.g., \cite{palacios2016distributed, morgan2016swarm, rathnam2013distributed}] exploit application-specific dynamics, therefore their solutions cannot be generalized to a broader context. In other words, if the problem objectives or the dynamics are changed, most of the existing approaches must be redesigned from scratch to adequately tackle the altered problem.

	\subsection{Contributions}
	
	To overcome the aforementioned problems, we propose a new resource optimization algorithm, specifically tailored to the context of multi-robot applications, that extends the cognitive-based adaptive optimization (CAO) algorithm \cite[]{kosmatopoulos2009adaptive}. CAO was originally developed and analyzed for the optimization of functions for which an explicit form is unknown but their measurements are available, as well as for the adaptive fine-tuning of large-scale nonlinear control systems \cite[]{kouvelas2011adaptive, kosmatopoulos2009large}.
	
	In a nutshell, an update cycle on decision variables of the proposed algorithm consists of the following steps. Initially, the robots' measurements are gathered in a central node (robot or base station) where the calculation of the global objective function takes place. In the following, each robot's contribution to the cost function is approximated and forwarded to the corresponding robot. In a fully distributed fashion, each robot constructs a linear-in-the-parameters (LIP) estimator to approximate the (unknown, problem-dependent) evolution of its subcost function. Then, each robot generates random (or pseudo-random) perturbations around its current state and neglects those that violate the operational constraints (if any). Finally, the next robot's action is the one valid perturbation that achieves the best score on the previously constructed estimator. 
	
	The proposed algorithm deviates from the original version of CAO in its distributed nature. More precisely, although each robot does not know explicitly either the decision variables of the other robots nor of their measurements, it is able to update its own decision variables effectively in a way to cooperatively achieve the team objectives. The latter can be achieved through a cost function that is exclusive to each robot, designed so as to encapsulate not only the mission objectives but also the other robots' dynamics (``data-driven gradient descent'' approach: for more details see Section \ref{sec:proposedAlg}). Rigorous arguments establish that despite the fact that the dynamics that govern the multi-robot system are unknown, the proposed methodology shares the same convergence characteristics as those of block coordinate descent algorithms \cite[]{wright2015coordinate}. As exhibited in the presented applications, the distributed nature of the proposed algorithm also allows rapid convergence, especially in cases with many robots. 
	
	The contributions with respect to the multi-robot approaches as presented in the previous subsection are as follows.
	\begin{itemize}
	\item[(i)]  The problem is formulated in a continuous domain without the need to either know all the states and measurements beforehand, or to perform a relaxation on the original multi-robot problem  (\textit{optimal control} and \textit{dynamic programming} approaches). The ability to cope with unknown dynamics (robots--environment) and unknown cost functions imparts a generality to the proposed algorithm, regarding the spectrum of applications that can be utilized.
	
	\item[(ii)]  However, the main advantage of the proposed algorithm is that it does not require either a priori calculation of the cost function (\textit{optimal one-step-ahead} approaches) or the analytical form of the system to be optimized to be explicitly known (\textit{optimal control} and \textit{dynamic programming} approaches). Instead, the proposed algorithm can cope with cost functions whose calculation can only be achieved by actually performing the corresponding course of actions. Along the same lines, the proposed algorithm does not require evaluation of the decision variables in the vicinity of their current values for calculating their corresponding updates. Instead, the proposed algorithm is able to find the (locally) optimal configuration for the decision variables by using only noise-corrupted measurements collected from the robots' sensors.
	
	\item[(iii)]  Furthermore, instead of relying on exhaustive, computationally intensive simulations (\textit{simulation-based} approaches), the proposed scheme is able to online learn the problem-specific characteristics that affect the user-defined objectives. By doing so, the proposed algorithm does not need any elaborate model in order to learn its decision-making mechanism.
	\end{itemize}
	
	It must be emphasized that apart from rendering the optimization problem practically solvable, the proposed approach preserves additional features that make it particularly tractable:
	\begin{itemize}
		\item[(i)] its complexity is low, allowing \textit{real-time implementations};
		\item[(ii)] it can handle a variety of \textit{physical constraints};
		\item[(iii)] it has \textit{fault-tolerant characteristics}, i.e., online redesign in case one or more robots being added or removed, an extra task being added to the set of objectives, etc.;
		\item[(iv)] it is able to adapt its behavior even in cases where a \textit{time-varying objective function} is employed\endnote{The rate of change in the objective function should be smaller than the learning capabilities of the algorithm (see Section \ref{sec:problemForm})}.
	\end{itemize}
	
	\subsection{Simulation testbeds}\color{black}
	The proposed control strategy is evaluated on \textit{three different simulation set-ups} under multiple scenarios, against both general-purpose and problem-specific algorithms. All the simulation set-ups have been chosen so that: i) the objective of the multi-robot mission can be expressed as a cost function, and ii) the evaluation of which cannot be performed beforehand. 
	
	In the first simulation set-up, the objective is to spread out the robots over a 2D environment while aggregating in areas of high sensory interest. An important aspect of the set-up is that the robots are not aware beforehand of the sensory areas of interest - instead, they learn this information online via sensor measurements from their current positions. The proposed algorithm is evaluated together with the approach proposed by \cite{schwager2009decentralized} for the problem in hand. 
	
	In the second simulation set-up, the trajectories of the robots should be designed in real-time having a twofold objective (which forms a trade-off). On the one hand, the part of the 3D terrain that is monitored (i.e., visible) by the robots has to be maximized and, on the other hand, for each one of these visible points in the terrain, the closest robot has to be as close as possible to that point. This problem along with a centralized CAO-based methodology has been proposed by \cite{renzaglia2012multi}, therefore a detailed analysis regarding the performance of both algorithms, in different scenarios, is presented. 
	
	Last but not least, the proposed methodology is evaluated in the task of persistent coverage. The objective of this application is to maintain a user-defined level of coverage in an unknown environment \cite[]{palacios2016distributed}. This is a quite challenging task as the mission objectives constantly change, whereas the unknown morphology of the environment does not allow the prior calculation of the improvement in the coverage task. 
	
	Conclusively, if it is possible to define a cost function which encapsulates the mission objectives and can be calculated through the robots' measurements for every decision variables configuration, the proposed methodology will be directly applicable to the corresponding problem.
	
	\subsection{Paper structure}\color{black}
	The remainder of the paper is structured as follows. Section \ref{sec:problemForm} presents the translation of a general-purpose multi-robot framework to a constrained optimization problem, highlighting the difficulties and the obstacles of the general problem. The description of the proposed algorithm, which tackles such a problem, is presented in Section \ref{sec:proposedAlg}. Sections \ref{sec:2dCoverage}, \ref{sec:3Dcoverage}, and \ref{sec:persistent} present three indicative multi-robot applications: \textit{adaptive coverage of unknown 2D environment}, \textit{3D surveillance of unmapped terrains}, and\textit{ persistent coverage of unknown 2D environments}, respectively. In all these sections, we perform a series of simulations in different scenarios to adequately analyze the performance of the proposed algorithm. The overall conclusions of the paper are drawn in Section \ref{sec:conclusions}.

	\section{Problem formulation}	
	\label{sec:problemForm}

	Consider a team (swarm) that consists of $N$ robots interacting with each other, towards achieving a global set of objectives. Let us assume the following augmented decision vector
	\begin{equation}
		\label{eq:stataSpace}
		\textbf{x}(k) \equiv \left[x_1^\tau(k), x_2^\tau(k),\dots,x_N^\tau(k) \right]^\tau 
	\end{equation}	
	where $x_i(k) \in \mathbb{R}^n$ denotes the decision variables of the $i$th robot at the $k$ th iteration. These decision variables represent the controllable parameters of the available robots (e.g., position, motors, propellers, thrusters, rotation of the cameras, etc.). Furthermore, the augmented vector which contains the available exteroceptive measurements takes the form
	\begin{equation}
		\label{eq:measurements}
		\textbf{y}(k) \equiv \left[y_1^\tau(k), y_2^\tau(k),\dots,y_N^\tau(k) \right]^\tau 
	\end{equation}
	where $y_i(k) \in \mathbb{R}^m$ denotes the measurement vector of the $i$th robot at the $k$ th iteration and its evolution can be represented as
	\begin{equation}
		\label{eq:measurmentsUpdate}
		y_i(k) \equiv h_i(k,x_i(k))
	\end{equation}
	where $h_i(\cdot)$ denotes an unknown, nonlinear function that depends on both $x_i(k)$ and the specific problem characteristics. 
	
	The accomplishment of the multi-robot system's objectives (e.g., mapping, surveillance, etc.) can be translated into the minimization (or maximization)\endnote{Without loss of generality, in the rest of the paper we assume a minimization problem.} of a specifically defined global cost function $\jmath_k$, i.e.,
	
	\begin{equation}
		\label{eq:globalCFx}
		\jmath_k \equiv {\cal J} \bigg(x_1(k), x_2(k),\dots,x_N(k) \bigg)
	\end{equation}	
	
	where ${\cal J}(\cdot)$ is a non-negative, nonlinear, scalar function that depends, apart from the robots decision variables, on the particular dynamics of the problem (e.g., the environment where the robots operate). Owing to the dependence of the function ${\cal J}$ on the particular problem characteristics, the \textit{explicit form of the function ${\cal J}$ is not known} in practical scenarios; as a result, standard optimization algorithms (e.g.,   gradient descent  with an a priori model) are not applicable. However, in most practical cases, the current value of the objective function can be approximated from the robots' measurements,
	
	\begin{equation}
		\label{eq:globalCFy}
		{\cal J} \bigg(x_1(k),\dots,x_N(k) \bigg) = \mbox{J} \bigg(y_1(k), \dots,y_N(k) \bigg) + \xi_k
	\end{equation}	
	
	where $\xi_k$ denotes the noise introduced in the estimation of $\jmath_k$, owing to the presence of noise in the robots sensors.\endnote{Note that, although it is natural to assume that the noise sequence $\xi_k$, is a stochastic zero-mean signal, it is not realistic to assume that it satisfies the typical AWNG property, even if the robots sensors do; as ${\cal J}$ is a nonlinear function of the robots decision variables and, thus, of the robots sensor measurements (\ref{eq:measurmentsUpdate}), the AWNG property is typically lost.} It must be emphasized that, in contrast to ${\cal J}$, $\mbox{J}$ can be evaluated ``offline'', if the measurement vector $\textbf{y}(k)$ is available. However, the \textit{acquisition of a new measurement vector} requires an \textit{actual evaluation} of the decision variables on the robotic system (\ref{eq:measurements}) and (\ref{eq:measurmentsUpdate}). 
	
	Apart from the problem of dealing with a criterion for which an explicit form is not known, but only its noisy measurements are available at each time, the decision vector $\textbf{x}(k)$ should satisfy a set of constraints that, in general, can be represented as follows:
	\begin{equation}
		\label{eq:constraints}
		{\cal C} \left( \textbf{x}(k) \right)  \leq 0
	\end{equation}
	
	where ${\cal C}$ is a set of nonlinear functions of the decision variables $\textbf{x}(k)$. As in the case of ${\cal J}$, the constraints function ${\cal C}$ depends on the particular problem characteristics and an explicit form of this function may be not known in many practical set-ups; however, it is natural to assume that the low-level algorithm is provided with information whether a particular selection of decision variables $\textbf{x}(k)$ satisfies or violates the set of constraints (\ref{eq:constraints}).
	
	Given the mathematical description presented above, the problem of choosing the decision variables online for a multi-robot system, so as to accomplish a set of objectives, can be mathematically described as the following constrained optimization problem:
	\begin{equation}
		\label{eq:OptimizationProblem}
		\begin{array}{rl}
			\mbox{minimize} & \jmath_{k}\\
			\mbox{subject to} & {\cal C}\left(\textbf{x}(k)\right) \leq 0 \,
		\end{array}
	\end{equation}
	
	As already noted, the difficulty in solving the constrained optimization problem (\ref{eq:OptimizationProblem}) in real-time lies in the fact that explicit forms for the functions ${\cal J}$ and ${\cal C}$ are not available. Although this is not the only problem, jointly optimizing a function over multiple robots ($N$), each of which with multiple decision variables ($n$), can incur excessively high computational cost.
	
	\section{Proposed algorithm}
	\label{sec:proposedAlg}

		\begin{figure*}[!th]
		\centering
		\includegraphics[width=0.99\textwidth]{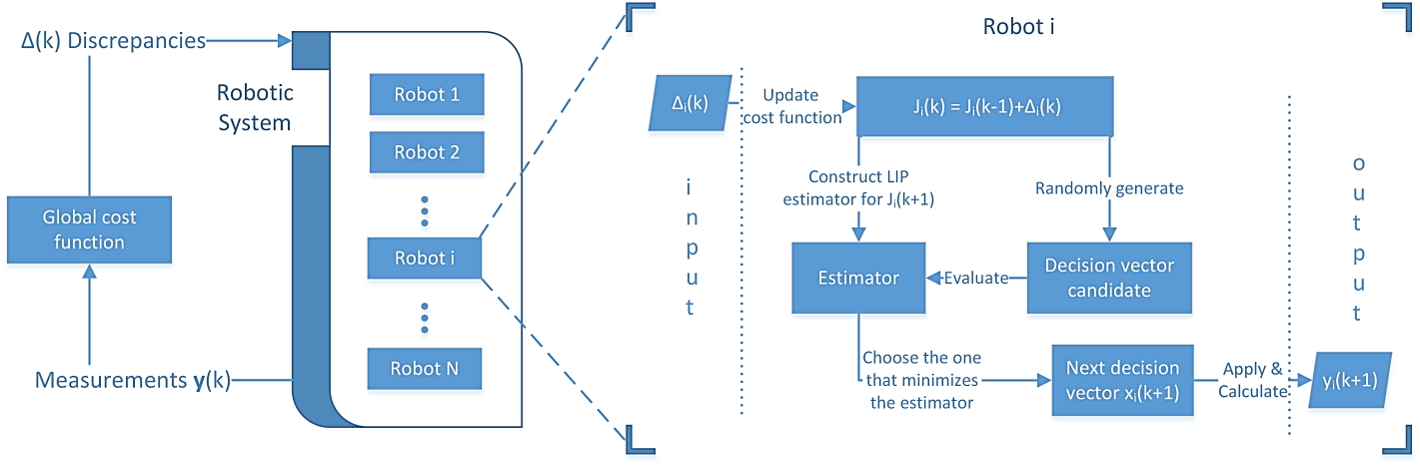}
		\caption{High-level diagram of the proposed algorithm. At each timestamp, all the operational robots first apply their decision commands and acquire the corresponding measurements $\textbf{y}(k)$, in order to be able to calculate the \textit{global cost function} index. Then, the contribution $\Delta_i(k)$ of the each robot to overall accomplishment of mission objectives is calculated and sent to the $i$th robot. In a fully distributed fashion, each robot constructs a linear-in-the-parameters estimator to approximate the (unknown - problem dependent) evolution of its sub-cost function $J_i(k)$, which encapsulates both the mission objectives and the operational capabilities of the multi-robot team. Finally, each robot’s next decision vector $x_i(k+1)$ is the one valid perturbation that achieves the best score on the previously constructed estimator.}
		\label{fig:highLevelDiagram}
	\end{figure*}
	
	Having defined the fundamental aspects that govern a multi-robot application, we proceed to present the proposed algorithm for updating the decision variables $\textbf{x}(k)$ so as to minimize the cost function (\ref{eq:globalCFx}) subject to (\ref{eq:constraints}).  A high-level diagram of the proposed algorithm is sketched in Figure \ref{fig:highLevelDiagram}. 
	
	\subsection{Global coordination}
	\label{subsec:globalCoordination}
	\textit{Step 1}. As a first step, and for each iteration $k$, the robots transmit the acquired measurements, after the execution of $\textbf{x}(k)$ decision variables. 
	
	It must be emphasized that this step can be performed even in cases where global communication between all robots is not feasible. In such a case, each robot can send and receive measurements to and from peer (adjacent) robots, until all the measurements aggregate to the corresponding processor unit (robot or ground station). The latter can be guaranteed by introducing an extra condition on the constraints set (\ref{eq:constraints}), ensuring the connectivity, if applicable to the problem in hand, among the different robots. 
	
	\textit{Step 2}. Thus, the global cost function can be straightforwardly derived from (see (\ref{eq:globalCFx}) and (\ref{eq:globalCFy})):
	$$
	\jmath_{k} = \mbox{J} \bigg(y_1(k), \dots,y_N(k) \bigg) 
	$$
	In addition, for each $i$ th robot calculate, the following discrepancy: 
	\begin{equation}
		\label{eq:subCF}
		\begin{aligned}
			&\Delta_i(k)  \equiv \jmath_{k} - \\
			& \mbox{J} \bigg(y_1(k),\dots,y_{i-1}(k),y_i(k-1),y_{i+1}(k)\dots,y_N(k) \bigg) 
		\end{aligned}
	\end{equation}
	
	In other words, $\Delta_i(k)$ encapsulates the effect of the $x_i(k)$ on the current problem for the $k$th timestamp. 
	
	Note that, because the last term of (\ref{eq:subCF}) is analytically available, we can calculate this term, although the resulting value does not necessary correspond to the actual value when the robots have the following decision variables:
	$$
	\left\lbrace x_1(k),\dots,x_{i-1}(k),x_i(k-1),x_{i+1}(k)\dots,x_N(k)\right\rbrace
	$$
	Although, there may be a discrepancy between the way we calculate $J(\cdot)$ and its actual value, that does not affect the convergence properties of the proposed algorithm. This discrepancy is application oriented and depicts the effect of other robots’ decisions on each robot's measurements. If the measurements acquired from a robot only affects its own decision variables and the problem itself (\ref{eq:measurmentsUpdate}), then there is no discrepancy at all.
	
	
	
	
	
	\textit{Step 3}. Next, the calculated discrepancy $\Delta_i(k)$ is sent to the i $th$ robot. 
	
	After this step all the calculations are performed locally, building a system that i) is resilient to robot failures, ii) does not require any global coordination, and ii) all the decision variables' updates are made in a (parallel) distributed fashion.
	
	\subsection{Distributed decision}
	\label{subsec:distributedDecision}
	Each $i$th robot, at the same $k$-th iteration, performs the following.
	\begin{enumerate}[label=(\alph*)]
	
	\item Calculate the $J_i(k)$ that corresponds to the last executed decision variables $x_i(k)$ as
	\begin{equation}
		\label{eq:constituentCF}
		J_i(k) = J_i(k-1) + \Delta_i(k) , \; \forall k\ge 1, \;J_i(0) = \jmath_0
	\end{equation} 
	
	Therefore, each robot is responsible to choose the next values for its decision variables $x_i(k+1)$, having as only objective the minimization of its corresponding cost function $J_i(\cdot)$\endnote{In general case: $\sum_{i=1}^{N}J_i(k) \ne \jmath_k$ and $\prod_{i=1}^{N} J_i(k) \ne \jmath_k, \; \forall k$}. Each such sub-problem is a lower-dimensional minimization problem, and thus can typically be solved more easily than the full problem.
	
	\item Construct a LIP estimator of $J_i(k+1)$ as follows:
	\begin{equation}
		\label{eq:estimatorJ}
		J_i(k+1) \approx \hat{J}_i(k+1) = \theta_i^\tau(k)\phi_i\big(x_i(k)\big) 
	\end{equation}
	
	where $\phi_i$ denotes the nonlinear vector of $L$ \textit{regressor terms}, $\theta_i$ denotes the vector of the \textit{parameter estimates} and $L$ is a positive user-defined integer which denotes the size of the function approximator.
	Defining the vector of regressor terms $\phi_i$ as in Section \ref{subsubsec:monomials_construnction}, the estimator vector $\theta_i$ can be calculated using standard least-squares estimator principles, i.e., $\theta_i$ is obtained by solving the following optimization problem:
	
	\begin{equation}
		\label{eq:estimatorTH}
		\theta_i(k) = \argmin_\vartheta \sum_{\ell = k-T(k)}^{k-1} \bigg( \vartheta^\tau \phi_i\big(x_i(\ell)\big) - J_i(\ell+1)\bigg)^2
	\end{equation}
	where $T(k)$ denotes the time window over which the least-squares estimation is taking place.
	
	\item Generate (randomly or pseudo-randomly) a set of $M$ valid candidate perturbations: $$\delta x_i^{(1)}(k),\delta x_i^{(2)}(k),\dots,\delta x_i^{(M)}(k)$$
	where $\delta x_i^{(j)}(k)$  are vectors of the same dimension as $x_i(k)$ and $M$ is a positive integer that is larger\endnote{See \cite{kosmatopoulos2009adaptive} for more details about the sufficiency of this condition.} than 2n. A candidate perturbation $j$ is considered valid if\endnote{The distributed nature of the algorithm may impose a stricter set of constraints, in comparison with cases where a centralized control is applied.} :
	\begin{equation}
		\label{eq:Distributedconstraints}
		{\cal C} \left(   \left[x_1^\tau,  \dots, x_{i-1}^\tau,  x_i^\tau  + \delta x_i^{(j)}, x_{i+1}^\tau,   \dots,  x_N^\tau   \right]^\tau \right)  \leq 0
	\end{equation}
	
	The random choice for the candidates is essential and crucial for the efficiency of the algorithm, as such a choice guarantees that $\hat{J}_i(k+1)$ is a reliable and accurate estimate for ${J}_i(k+1)$; see \cite{kosmatopoulos2009adaptive} and \cite{kosmatopoulos2009large} for more details.

	\item Estimate the effect of each of the candidate perturbations on the current vector $x_i(k)$ by employing the previously constructed estimator (\ref{eq:estimatorJ}) and pick the candidate perturbation with the ``best'' effect, i.e., choose the vector $\delta x_i^{(j^*)}(k)$ that satisfies
	
	$$
	\delta x_i^{(j^*)}(k) = \argmin_{j=1,\dots,M} \theta_i^\tau(k)\phi_i\bigg(x_i(k) + \alpha(k)\delta x_i^{(j)}(k)\bigg)
	$$
	
	\item Update the $i$ th robot decision variables as
	
	\begin{equation}
		\label{eq:updateRule}
		x_i(k+1) = x_i(k) + \alpha(k)\delta x_i^{(j^*)}(k)
	\end{equation}
   where $\alpha(k)$ is a positive function chosen to be either a constant positive function or a time-descending function satisfying $\alpha(k) >0, \sum_{k=0}^\infty \alpha(k) = \infty, \sum_{k=0}^\infty \alpha(k)^2 < \infty$.  Furthermore, $\alpha(k) \leq \bar{\alpha} \; \forall k$, where $\bar{\alpha}$ is a problem-specific constant, correlated with the robot's dynamics (e.g., maximum achievable movement in one timestamp) and the objectives of the multi-robot application. 
   
   \item Finally, by applying the $x_i(k+1)$ decision vector, the corresponding $y_i(k+1)$ measurements vector will be acquired. This vector, along with all the measurements from the remaining robots, are utilized in order to evaluate the $k+1$ team configuration (see \textit{Step 1} from the previous subsection).
   \color{black}
\end{enumerate}

	\begin{remark}
		\label{rm:1}
		\normalfont
		The above distributed update of the decision variables (Section \ref{subsec:distributedDecision}) does not need information about what is happening to the other robots. All the necessary information has been ``packed'' to the scalar value $\Delta_i(k)$. At each iteration, each robot attempts to minimize the objective function $J_i(k)$ by assuming that the other robots' decision variables are part of the problem to be solved.
	\end{remark}

	\begin{remark}
	\label{rm:localMinima}
	\normalfont
	The utilization of random perturbations provides the proposed algorithm with the potential to escape from local minima. In essence, the random perturbations inside the distributed decision mechanism (step (c)), could have a behavior similar to \textit{simulated annealing}, which has been proved that under specific conditions can overcome local minima \cite[]{granville1994simulated} that may arise from the distributed nature of the algorithm.
	\end{remark}
	\color{black}

	\subsection{Estimator implementation}
	\label{subsec:estimator_implementation}
	This subsection encloses the implementation details of the $i$th robot estimator (\ref{eq:estimatorJ}), as outlined in step (b) of the distributed decision-making scheme.
	\subsubsection{$\phi$ monomial construction.}
	\label{subsubsec:monomials_construnction}
	The vector $\phi_i$ of regressor terms must be chosen so that it satisfies the so-called \textit{Universal Approximation Property} \cite[]{polycarpou1991identification}, i.e., it must be chosen so that the approximation accuracy of the constructed approximator (\ref{eq:estimatorJ}) is an increasing function of the approximator's size $L$. Polynomial approximators, radial basis functions, kernel-based approximators, etc. are known to satisfy such a property \cite[]{polycarpou1991identification}.\color{black}
	
	Experimenting with different types of $\phi_i$, in different multi-robot set-ups (Sections \ref{sec:2dCoverage}--\ref{sec:persistent}, and \cite{chatzichristofis2013autonomous, kapoutsis2015real}), it was found that it is sufficient to construct a polynomial estimator as in Algorithm \ref{alg:regressorVector}.
	
	\begin{algorithm}[h]
		\caption{$\phi_i$ construction}
		\label{alg:regressorVector}
		\begin{algorithmic}[1]
			\REQUIRE maxorder, $L_1,L_2,\dots,L_{\text{maxorder}}$, $ x_i$, $ n$
			\ENSURE $\phi_i$
			\STATE $\phi_i = 1 $
			\FOR{$j \in\{1,...,\text{maxorder}\}$}  
			\FOR{$v \in\{1,...,L_j\}$} 
			\STATE $g=1$
			\FOR {$l \in\{1,...,j\}$} 
			\STATE Generate $r:=$ random integer $ \in \{1,\dots,n\}$ 
			\STATE $g = g \cdot x_i^{(r)}$
			\ENDFOR
			\STATE $\phi_i = \left[\phi_i^\tau, g \right]^\tau $
			\ENDFOR
			\ENDFOR
		\end{algorithmic}
	\end{algorithm}
	
	The tunable parameters of this procedure are the maximum order of monomials (maxorder) and the corresponding number of monomials per order ($L_1,L_2,\dots,L_{\text{maxorder}}$, where $L_1 + L_2 + \dots + L_{\text{maxorder}} = L-1$ should be hold). Mathematically speaking, the number of different monomials per order is given by the number of possible combinations with repetitions (multiset coefficient):
	$$
	\multiset{n}{i} = {n + i -1 \choose i} = \frac{n(n+1)(n+2)\cdots(n+i-1)}{i!}
	$$
	where ${a \choose b}$ denotes the binomial coefficient. However, the summation $L_1 + L_2 + \dots + L_\text{maxoder}$ may exceed the number of available monomials $L-1$. A usual practice is to downscale the number of monomials as follows: 
	\begin{equation}
	\label{eq:monomials}
	L_i = \left[ {n + i -1 \choose i}s\right] 
	\end{equation}
	where $[\cdot]$ denotes the nearest integer and $s$ denotes the following scaling factor
	$$
	s = \frac{L-1}{\sum_{i=1}^{\text{maxorder}} {n + i -1 \choose i}}
	$$
	
	\subsubsection{Solving the least-squares problem.} It is worth pointing out that although the $\hat{J}_i(k+1)$ is evolving in a nonlinear fashion with respect to $x_i$, standard linear regression techniques can be utilized to find $\theta_i$, as (\ref{eq:estimatorJ}) is still linear in the parameters' vector. Therefore, the least-squares problem as defined in (\ref{eq:estimatorTH}) can be solved by several algorithms (normal equation, QR decomposition, SVD, etc.).  Although, singular value decomposition (SVD) is more computational intensive in comparison to other alternatives, we utilize this approach due to the fact that it is more numerical stable (e.g., when the problem is ill-conditioned) \cite[]{demmel1997applied}.
	\color{black}
	
	\subsection{Convergence analysis}
	\begin{remark} 
		\label{rm:2}
		\normalfont
		As shown in \cite{kosmatopoulos2009adaptive, kosmatopoulos2009large}, the distributed algorithm implemented in each robot (Section \ref{subsec:distributedDecision}) guarantees that
		If $M \geq 2 \times \text{dim}\left(x_i\right)$, the vector $\phi$ satisfies the universal approximation property and the functions $J_i$ and $C$ are either continuous or discontinuous with a finite number of discontinuities, then the update rule of $x_i$ (\ref{eq:updateRule}) is equivalent to
		$$
		x_i(k+1) = x_i(k) - A(k) \nabla_{x_i} J_i + \epsilon(k)
		$$
		where $A(k)$ is a positive definite matrix that depends on the choice of $\alpha$ (see step (e) of the distributed decision-making scheme) and $\nabla_{x_i} J_i$ denotes the gradient of $J_i$ with respect to the $x_i$ decision variables.
		In addition, $\epsilon(k)$ is a term that converges exponentially fast to zero with probability one. In simple words, the analysis of \cite{kosmatopoulos2009adaptive, kosmatopoulos2009large} establishes that the algorithm will converge to a local minimum of $J_i$.
	\end{remark}
	

	The following theorem describes the properties of the proposed methodology; as the proof of this theorem is along the same lines as in \cite[Proposition 2.7.1]{bertsekas1999nonlinear}, only a sketch of proof is provided. 
	
	\begin{theorem}
		\label{th:1}	
		The local convergence of the proposed algorithm can be guaranteed in the general case where the global cost function ${\cal J}$ and each robot's contribution $J_i$ are non-convex, non-smooth functions.\endnote{Moreover, recent studies imply that BCD methodologies can achieve global convergence even in cases where the global cost function (\ref{eq:globalCFx}) is non-convex but holds some properties. For example, in \cite{xu2013block} the authors established global convergence of the BCD algorithm in the general case where the global cost function ${\cal J}$ and each robot's contribution $J_i$ are non-convex functions, but the so-called Kurdyka-\L{}ojasiewicz (KL) property is satisfied.} 
	\end{theorem}
	\textit{Sketch of the proof:} By using Remark \ref{rm:2} (\textit{projected gradient-descent} on the minimization of $J_i$) and Equations (\ref{eq:subCF})--(\ref{eq:constituentCF}), we can establish that the distributed update on each robot is equivalent to 
	$$
	x_i = \argmin_{w} {\cal J}\left(x_1,\dots,x_{i-1},w,x_{i+1},\dots,x_{N}\right)  \\
	$$
	subject to (\ref{eq:Distributedconstraints}) and therefore, the proposed algorithm approximates the behavior of the block coordinate descent (BCD) \cite[Algorithm 1]{wright2015coordinate} family of approaches. Following the proof described in \cite[Proposition 2.7.1]{bertsekas1999nonlinear}, it is straightforward to see that if the minimum with respect to each block of variables is unique, then any accumulation point of the sequence $\{x(k)\}$ generated by the BCD methodology is also a stationary point.
	
	\subsection{Complexity}
	\begin{table}
		\centering
		\begin{tabular}{ c c c c}
			\hline 
			\textit{STEP} & Complexity & Practical & Comments\\ \hline 
			\textit{1-3} & $ \mathcal{O}\left( J\left( \textbf{y}\right)  \right) $& $\mathcal{O}\left( N^3 m^2\right) $ & Application \\
			&&& dependent \\ 
			\textit{4} & $ \mathcal{O}\left( L^{3} \right)  $ & $\mathcal{O}(n^3)$ & Least-squares\\
			\hline 
		\end{tabular}
		\caption{Complexity analysis}
		\label{tbl:complexity}
	\end{table}

	The computational burden regarding the global coordination (section \ref{subsec:globalCoordination}) is accumulated in the calculation of $\Delta_i(k)$ (\ref{eq:subCF}) for each robot $i$. However, the calculation of J$\left(\cdot \right) $ is problem-dependent, thus, it is not possible to analytically derive bounds regarding its complexity. In the reported cases (cost functions (\ref{eq:slotiveCFj}),(\ref{eq:cf3dcover}),(\ref{eq:cf3dcoverTT}) and (\ref{eq:persistentCF})), as well as in most real-world applications, the computational needs of J$\left(\cdot \right) $ grow, at most, quadratic with the number of robots $\times$ the number of measurements per robot, i.e., $\mathcal{O}\left( N^2 m^2\right) $. Technically, the above threshold expresses the case where an operation is needed per different pair of measurements $\{ y_a^{(i)},  y_b^{(j)} \}$, with $a,b \in \{1, \dots,m\}$ and $i,j \in \{1, \dots, N\}$. Overall, J$\left(\cdot \right) $ is evaluated $N+1$ times, one for each robot and one for the global cost function term (\ref{eq:subCF}); therefore, \textit{Steps 1--3} are expected to have $\mathcal{O}\left( N^3 m^2\right) $. 
	
	The computational requirements for the distributed decision (Section \ref{subsec:distributedDecision}), which is computed on each robot, are dominated by the requirement of solving the least-squares problem (\ref{eq:estimatorTH}). According to \cite[Section 5.5.6, Figure 5.5.1]{golub2012matrix}, the best algorithms for least-squares problem using SVD procedure, take time that is proportional to $\mathcal{O}\left(T^2L + L^3 \right) $. In the interest of simplicity, and owing to the fact that $T \simeq L$, we can assume that the complexity for the distributed decision scales as $\mathcal{O}\left(L^3 \right) $. Although, there exist no theoretical results for providing the lower bound $\bar{L}$ for the size of the regressor vector, practical investigations on many different applications [e.g., \cite{korkas2016occupancy, kapoutsis2015real, amanatiadis2013multi}] indicate that it is sufficient enough to choose $L \ge \bar{L} = 2 \times n$, to adequately tackle the local approximation of $J_i$. Therefore, it is expected that the computational requirements will grow with $\mathcal{O}(n^3)$. Although this step is executed on each robot ($N$ times), the distributed nature of the algorithm guarantees that no extra computational needs will be required.
	
	Overall, it is expected that the complexity of computing $N$ times the cost function J$\left(\cdot \right) $ dominates the requirement of solving the least-squares problem for one robot. Table \ref{tbl:complexity} summarizes the complexity bounds discussed in this section. 
	
	\begin{remark}
		\label{rm:parameters}
		\normalfont
		We close this section by accumulating the free parameters of the proposed algorithm. The set is composed of the number of perturbations $M$, the total number of utilized monomials $L$ and the time window $T$ over which the least-squares estimation is taking place. According to Remark \ref{rm:2}, the number of perturbations $M$ should be greater than $2 \times n$. Furthermore, the complexity analysis of Section \ref{subsec:distributedDecision} indicates that the estimator (\ref{eq:estimatorJ}) should have at least $\bar{L} = 2 \times n$  number of monomials. Finally, $T$ is a non-negative integer that expresses the desired ``forgetting factor'' for the constructed estimator. In the following experimental set-ups, we set the algorithm's parameters within these bounds. Alternatively, and if required, all parameters mentioned could be manually tuned in order to achieve better, application-dependent, performance.
	\end{remark}

	\section{Adaptive coverage control utilizing Voronoi partitioning}
	\label{sec:2dCoverage}
	The first simulation set-up is the well-investigated optimal robots' placement problem \cite[]{schwager2009decentralized,schwager2006distributed,cortes2002coverage}. The objective for the network of robots is to spread out over an environment, while aggregating in areas of high sensory interest. Furthermore, the robots do not know beforehand where the areas of sensory interest are, but they learn this information online from sensor measurements. The aforementioned task can be found in applications such as environmental monitoring and clean-up, automatic surveillance of rooms/buildings/towns, or search and rescue missions.

	\subsection{Problem definition}
	
	It is assumed that the operational area is a bounded $Q \subset \mathbb{R}^n$. A point inside this environment is denoted by $q$ and the decision vector $x_i$ for the $i$ th robot contains its position in $Q$. In addition, let $\{V_1,\dots,V_N\}$ be the Voronoi partition of $Q$, for which the robot positions are the generator points:
	$$
	V_i = \{q\in Q | \left\| q-x_i\right\| \leq \left\| q-x_j\right\|, \forall j \neq i \}
	$$
	(Henceforth, we use $\left\| \cdot \right\| $ to denote the Euclidean norm $\left\| \cdot \right\|_2 $) Let $\zeta(\cdot)$ to be the unknown sensory function such that $\zeta : Q \rightarrow \mathbb{R}_{>0}$ (where $\mathbb{R}_{>0}$ is the set of strictly positive real numbers). In other words, this function $\zeta(\cdot)$ assigns in each location of the available space $Q$ a weight of importance related to the necessity of being covered.
	
	The global cost function for the problem in hand, admits the following form:
	\begin{equation}
		\label{eq:slotiveCF}
		\mathcal{J}(\textbf{x}(k)) = \sum^N_{i=1} \int_{V_i} \frac{1}{2}  \left\|q-x_i\right\|^2 \zeta(q) dq 
	\end{equation}
	Apparently, the above function cannot be calculated in advance owing to the dependence of the unknown sensory function $\zeta$. Without loss of generality, we assume that the sensory function is given by
	\begin{equation}
		\label{eq:approxPhi}
		\zeta(q) = \mathcal{K}(q)^\tau \upsilon + \mathcal{O}(1/W), \; \forall q \in Q
	\end{equation}
	
	where $ \mathcal{K}: Q \rightarrow \mathbb{R}_{>0}^W$  denotes a vector of bounded, continuous basis functions (e.g., Gaussians, wavelets, sigmoids, etc.) and $\upsilon \in \mathbb{R}^W$ is the parameter vector. The deviation from the actual value of $\zeta$ is in the order of the number of basis functions $ \mathcal{O}(1/W)$. Although $ \mathcal{K}$ is defined a priori, the mixing parameters vector $\upsilon$ is environment-dependent and generally unknown. However, the value of the sensory function can be measured from the robots' sensors (e.g., temperature/chemical sensor) at their current position's configuration $\textbf{x}(k)$. 
	\begin{equation}
		\label{eq:sensorModelSlotine}
		y(x_i) = \zeta(x_i) 
	\end{equation}
	
	The value of the parameter estimation vector $\hat{\upsilon}$ can be approximated through these measurements, utilizing standard parameter estimation techniques (e.g., least-squares approach (\ref{eq:estimatorTH})). Therefore, after the update on the parameter vector $\hat{\upsilon}$, a new update on the belief regarding the sensory function is also available through the equation
	\begin{equation}
		\label{eq:sesnoryFunctionUpdate}
		\hat{\zeta} = \mathcal{K}^\tau \hat{\upsilon}
	\end{equation}
	
	Hence, the value of the unknown cost function can be approximated through the following equation:
	\begin{equation}
		\label{eq:slotiveCFj}
		\mbox{J}(\textbf{y}(k)) = \sum^N_{i=1} \int_{V_i} \frac{1}{2}  \left\|q-x_i\right\|^2 \mathcal{K}^\tau(q) \hat{\upsilon} dq 
	\end{equation}

	\subsection{Simulation results}
	 For implementation reasons, we assume that the operation area consists of $225$ discrete points, uniformly distributed across the plane of $[0, 1]^2$. The sensory function, $\zeta(q)$, was parameterized as a linear combination of 49 Gaussians, i.e., $\mathcal{K}(j) = \frac{1}{2\pi\sigma^2_j} \mbox{exp} - \frac{(q-\mu_j)^2}{2\sigma^2_j}, \; \forall j \in \{1,\dots,49\}$. Each standard deviation is set to be $\sigma_j = 0.02$ and the Gaussians centers $\mu_j$ are chosen so as to be uniformly distributed in the operational area (seven Gaussians in each row and column).  The parameter $\upsilon = [\upsilon_1, \upsilon_2, \dots, \upsilon_{49}]^{\tau}$ was chosen so that $\upsilon_i=0.1, \forall i\in\{1,\dots,49\} $, apart from two random integers $a,b \in \{1,\dots,49\}$ whereas $\upsilon_a=\upsilon_b=100$. In other words, for each simulation instance, the sensory function $\zeta(q)$ was dominated by two, randomly selected, Gaussians. \color{black} Finally, the equations are integrated using a fixed step of $\alpha=dt=0.01$ and the initial values for the estimation of parameter vector (robots' knowledge) was chosen to be $\hat{\upsilon}=[0.1, 0.1, \dots, 0.1]^\tau$.
	
	
	In addition with the proposed approach, we present simulation results from the algorithm as proposed, for the problem in hand, by \cite{schwager2009decentralized}. The weights' selection was undertaken following the authors' instructions in \cite[Section 7.2]{schwager2009decentralized}. To construct comparable simulations instances, we utilize the same learning rule for the parameter vector $\hat{\upsilon}$ \cite[equation 13]{schwager2009decentralized}. In both the evaluated algorithms, the update of parameter vector was performed, by aggregating all the robots' measurements. To evaluate the performance of each approach in each timestamp, we also calculate the real value of the cost function (\ref{eq:slotiveCF}), but none of the evaluated algorithms utilizes this information.

	The proposed approach was employed with a constant time-window for the least-squares estimation of $T = 30$ and the number random perturbations was set to be $M=100$. To approximate each robot's cost function evolution, we utilize a third-order monomial estimator with $L = 10$ and using (\ref{eq:monomials}) we calculate the number of monomials per order to be $L_1 = 2, L_2=3 \text{ and } L_3 = 4$.

	\subsubsection{Random initial positions scenario.}
\begin{figure*}[!th]
		\centering
		\subfigure[]{\includegraphics[width=0.32\textwidth]{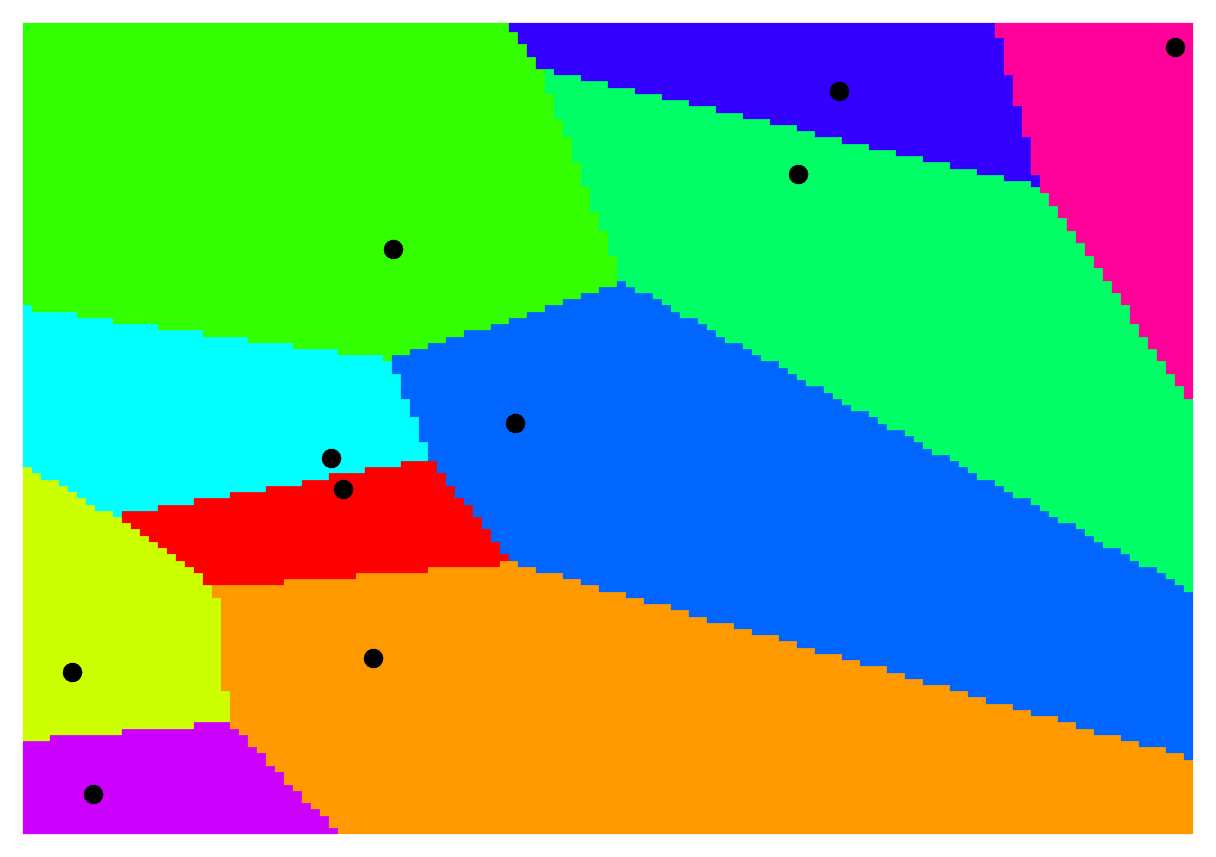}\label{fig:InitialVoronoiRandom}}
		\subfigure[]{\includegraphics[width=0.32\textwidth]{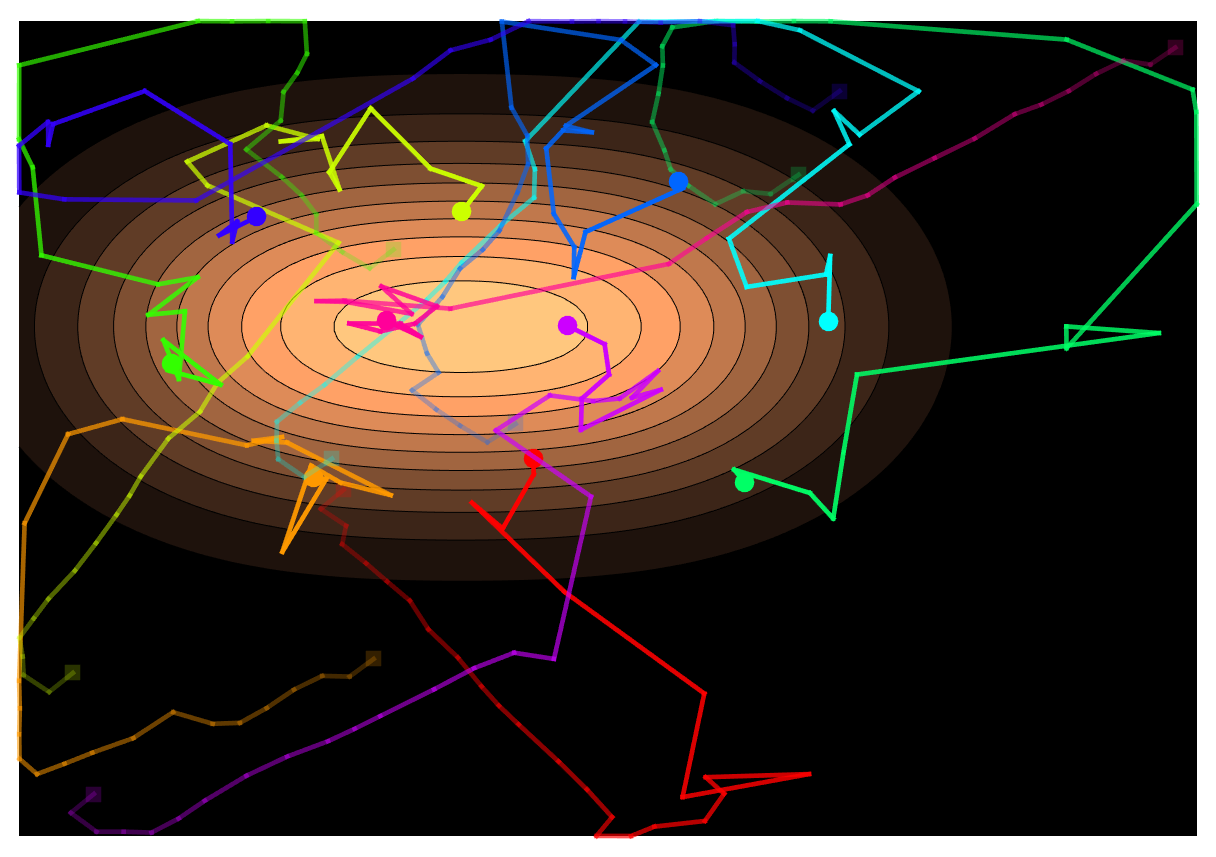}\label{fig:TrajectoriesRandom}}	
		\subfigure[]{\includegraphics[width=0.32\textwidth]{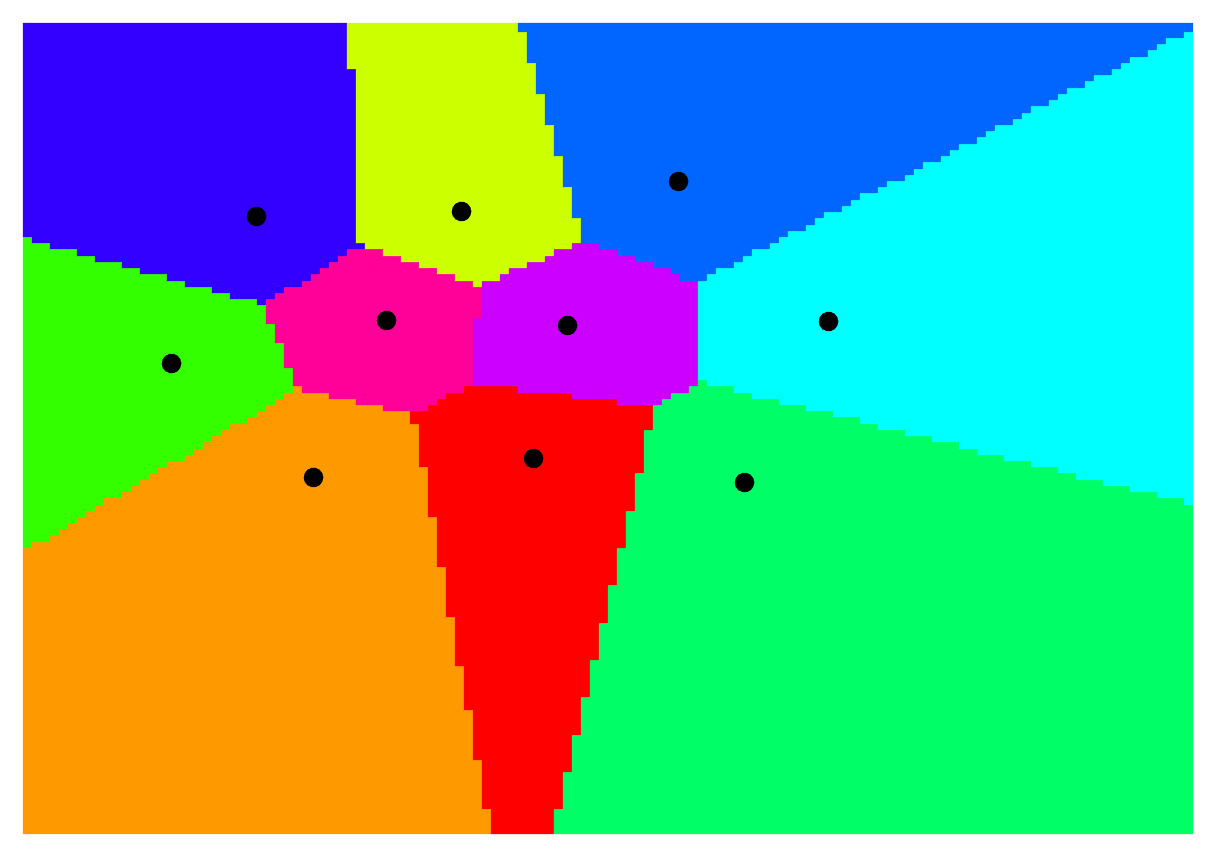}\label{fig:VoronoiRandom}}
		\caption{Illustrative example with random initial positions for the robots. (a) Initial Voronoi partitioning. (b) Robots’ trajectories on top of the heatmap of the sensory function. The squares and the circles denote the initial and the final positions of the robots, respectively.(c) Voronoi partitioning of the final configuration. In these figures, we sketch how the proposed algorithm drives the available robots so as to completely cover the space and to aggregate around areas with high sensory interest.}
		\label{fig:RandomScenario}
	\end{figure*}

	In the first simulation scenario, the robots were placed randomly along the x and y axes of the operation area.  An example of this simulation set-up is illustrated Figure \ref{fig:RandomScenario}, where Figure \ref{fig:InitialVoronoiRandom} sketches the Voronoi partitioning for the initial robot configuration. Figure \ref{fig:TrajectoriesRandom} illustrates  the robots' trajectories from their initial positions (squares) to the final configuration (circles), and, finally, Figure \ref{fig:VoronoiRandom} illustrates the Voronoi partitioning for the final robots' positions. As one can see, the robots gathered around the areas with the highest values of the unknown sensory function $\zeta(\cdot)$. \color{black}

\begin{figure*}[h]	
		\centering
		\subfigure[]{\includegraphics[width=0.47\textwidth]{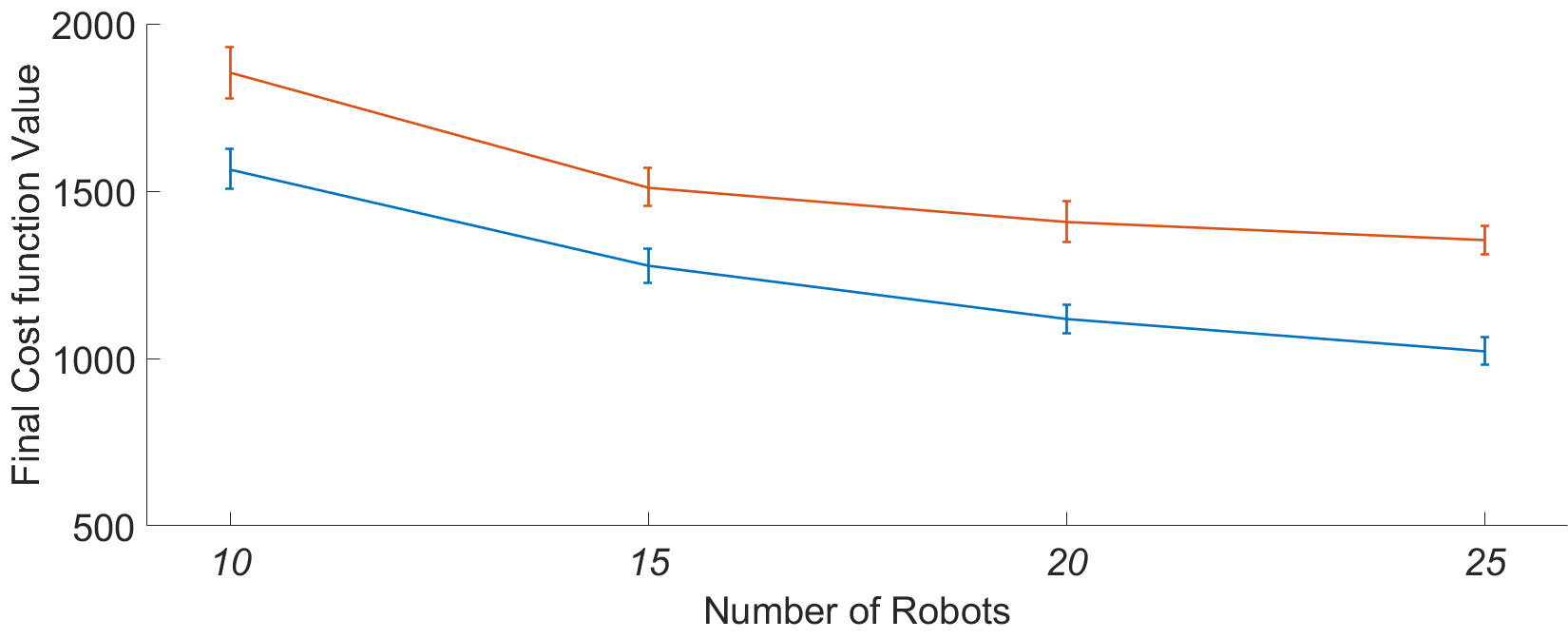}\label{fig:cfRandom}}
		\subfigure[]{\includegraphics[width=0.47\textwidth]{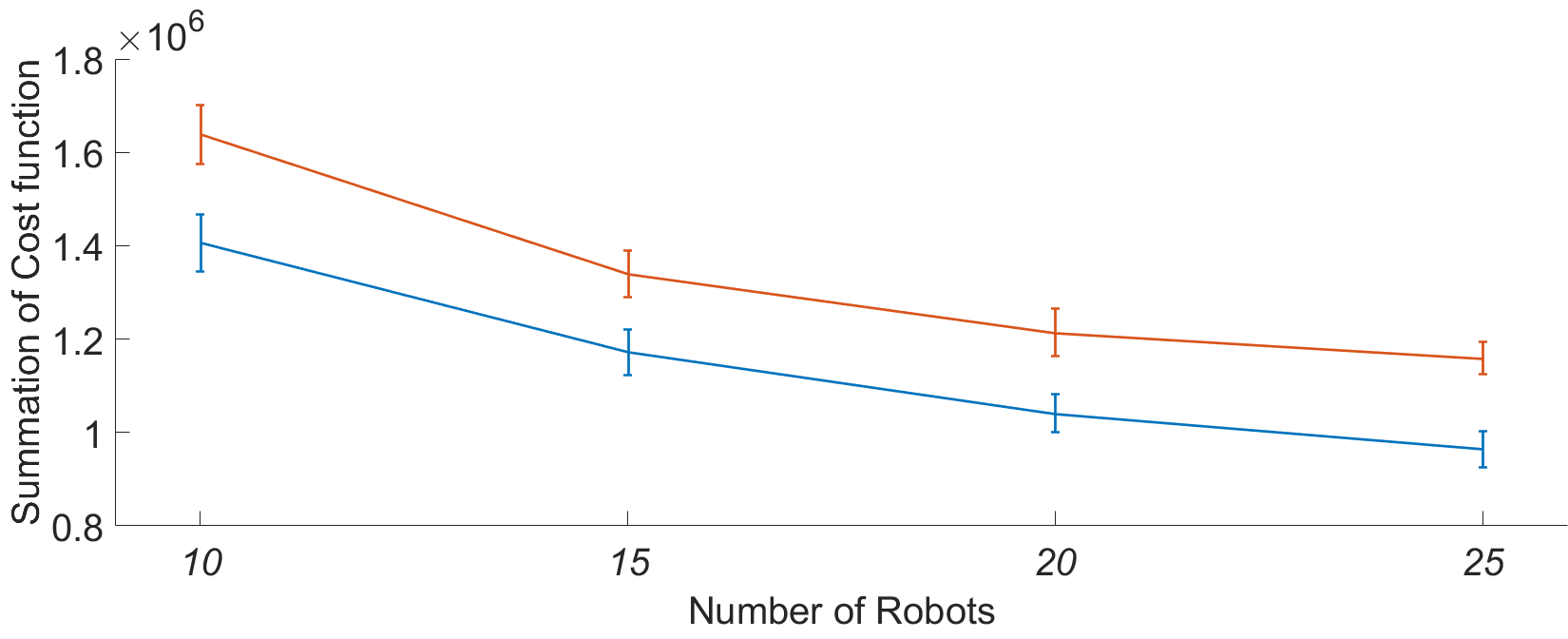}\label{fig:SumRandom}}
		\caption{Comparison study for the \textit{random initial positions scenario}: proposed algorithm (blue) and approach presented by \cite{schwager2009decentralized} (red). (a) Final achieved value of the cost function. (b) Summation of the cost function over the experiment’s horizon.}
		\label{fig:SlotineRandomScenario}
	\end{figure*}

	Figure \ref{fig:SlotineRandomScenario} presents a comparison study between the evaluated algorithms, over different sizes of robot teams. The number of robots was chosen to be 10, 15, 20, and 25 robots, and for each configuration 60 experiments with randomly selected initial robots' placement and sensory function were performed. The average, final achieved cost function (\ref{eq:slotiveCF}) values, along with the corresponding confidence intervals are illustrated in \color{black} Figure \ref{fig:cfRandom}.  In addition, we present the summation of the cost function over the course of each simulation pair (Figure \ref{fig:SumRandom}). It must be emphasized that, although the summation of the cost function may be strongly dependent on the initial robots' positions the final achieved value has a small variance around the average value. This feature highlights the ability of the proposed approach to converge to an optimal configuration, independently of the initial conditions. 
	
	\subsubsection{Right half-plane scenario.} 
\begin{figure*}[!th]
		\centering
		\subfigure[]{\includegraphics[width=0.32\textwidth]{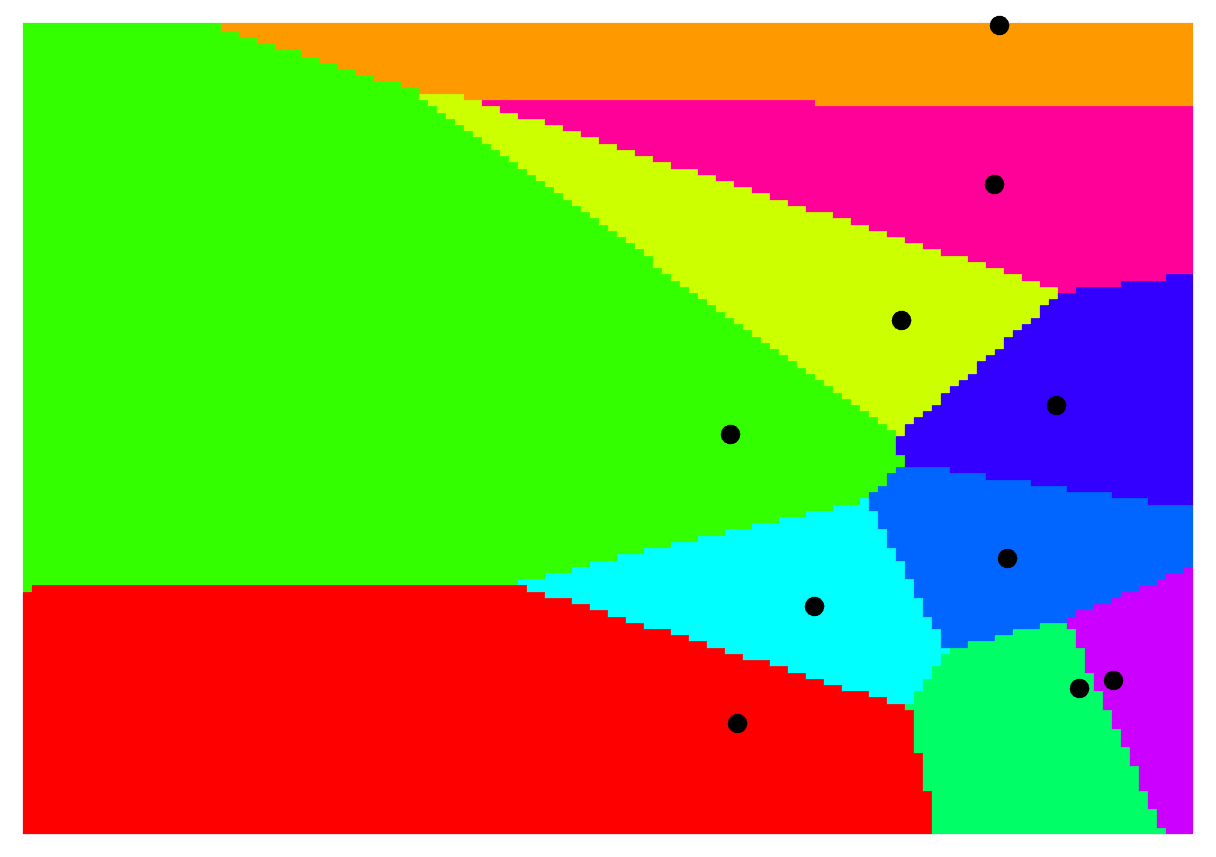}\label{fig:TrajectoriesHalfX}}
		\subfigure[]{\includegraphics[width=0.32\textwidth]{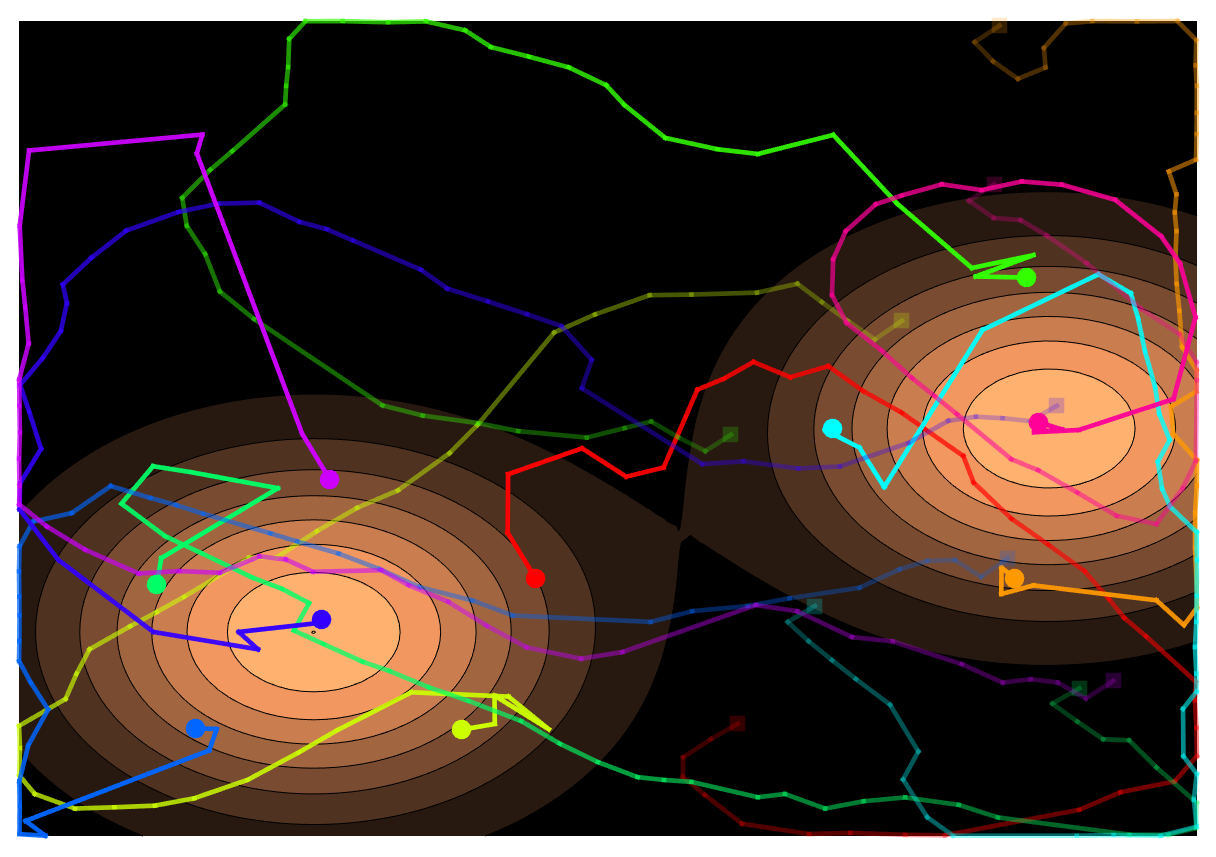}\label{fig:TrajectoriesHalfX}}
		\subfigure[]{\includegraphics[width=0.32\textwidth]{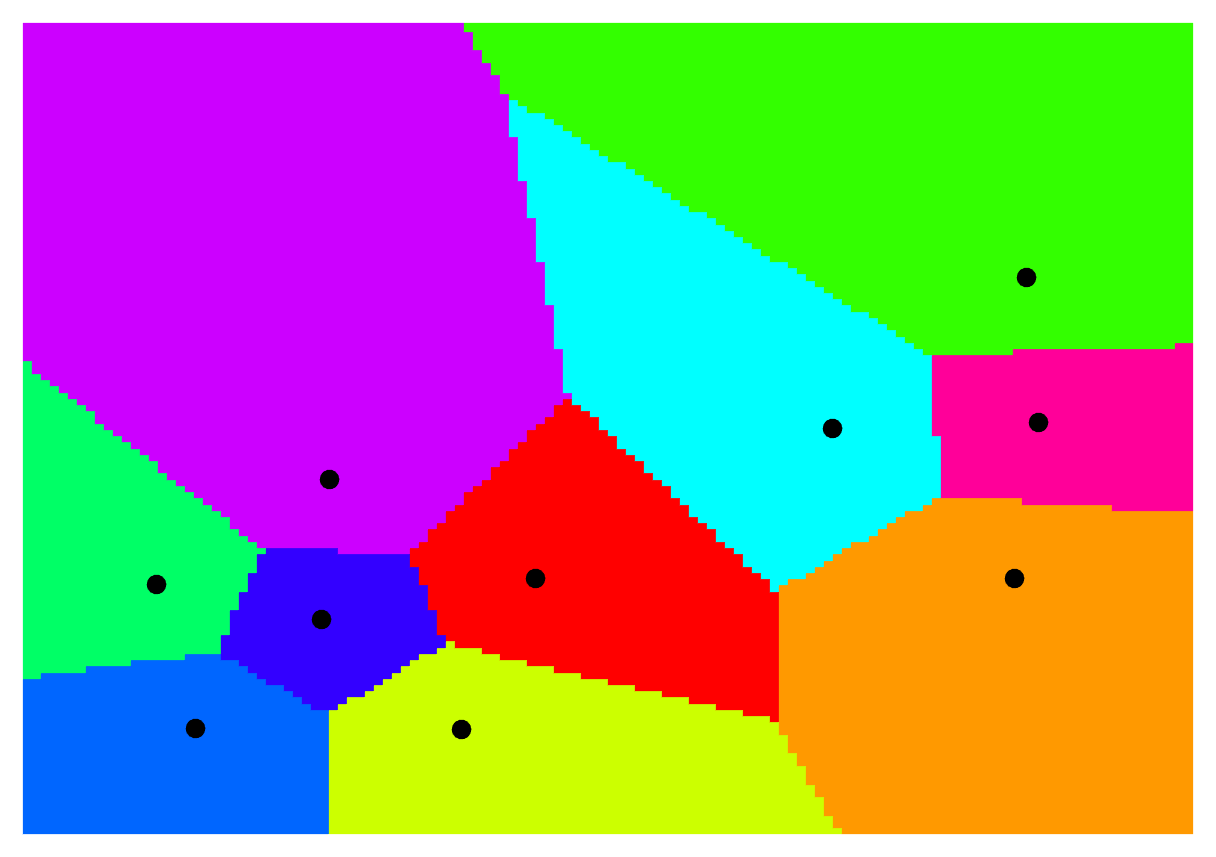}\label{fig:VoronoiHalfX}}
		\caption{Illustrative example where the robots initial positions are constrained inside the right half-plane of the operational environment. The proposed algorithm navigates the robots around the space, utilizing only their measurements on their current positions, to achieve the mission objective. (a) Initial Voronoi partitioning. (b) Robots’ trajectories on top of the heatmap of the sensory function. (c) Voronoi partitioning of the final configuration.}
		\label{fig:RightHalfPlaneScenario}
	\end{figure*}

In the second simulation scenario, the robots' initial positions were constrained inside the right half-plane of the operation area. In general, this scenario has a greater level of difficulty, compared with random initialization, as the robots can easily get stuck in highly suboptimal situations. Figure \ref{fig:RightHalfPlaneScenario} illustrates an instance of such a scenario where the proposed approach was utilized.
	
	\begin{figure*}[h]
		\centering
		\subfigure[]{\includegraphics[width=0.47\textwidth]{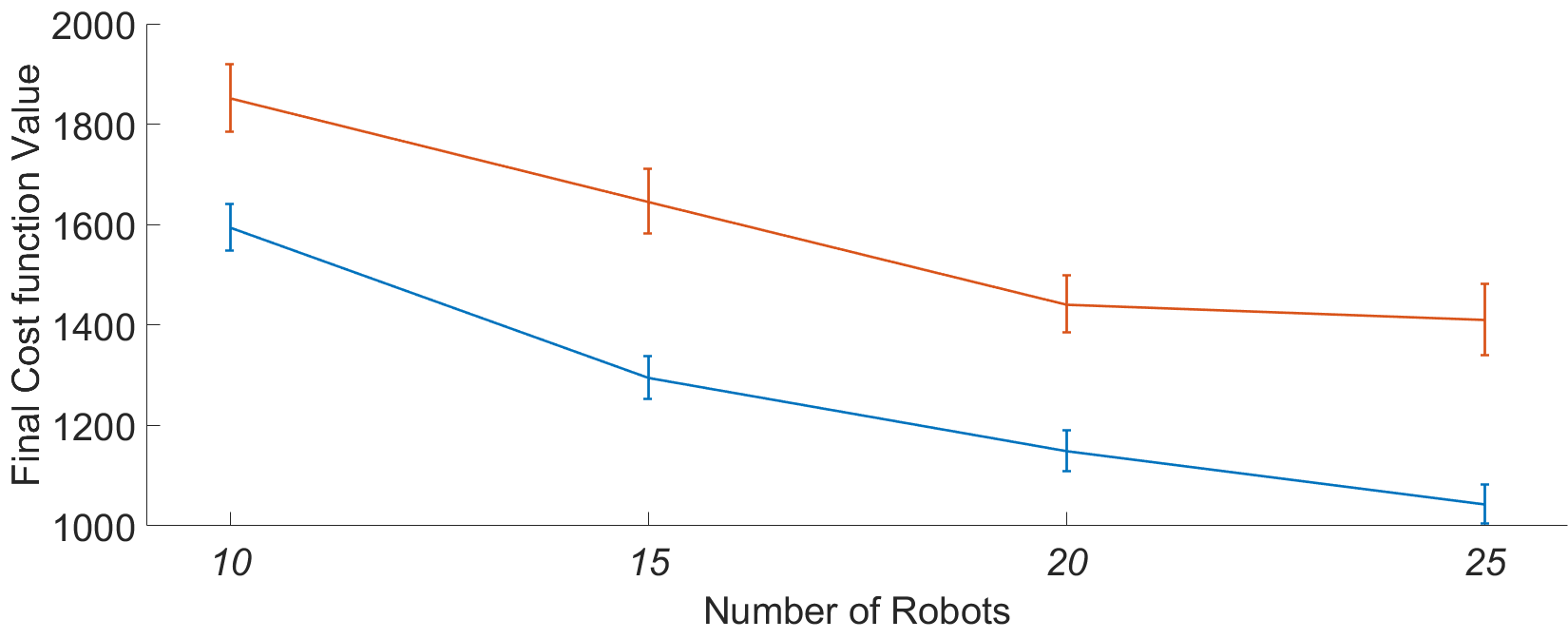}\label{fig:cfHalfX}}
		\subfigure[]{\includegraphics[width=0.47\textwidth]{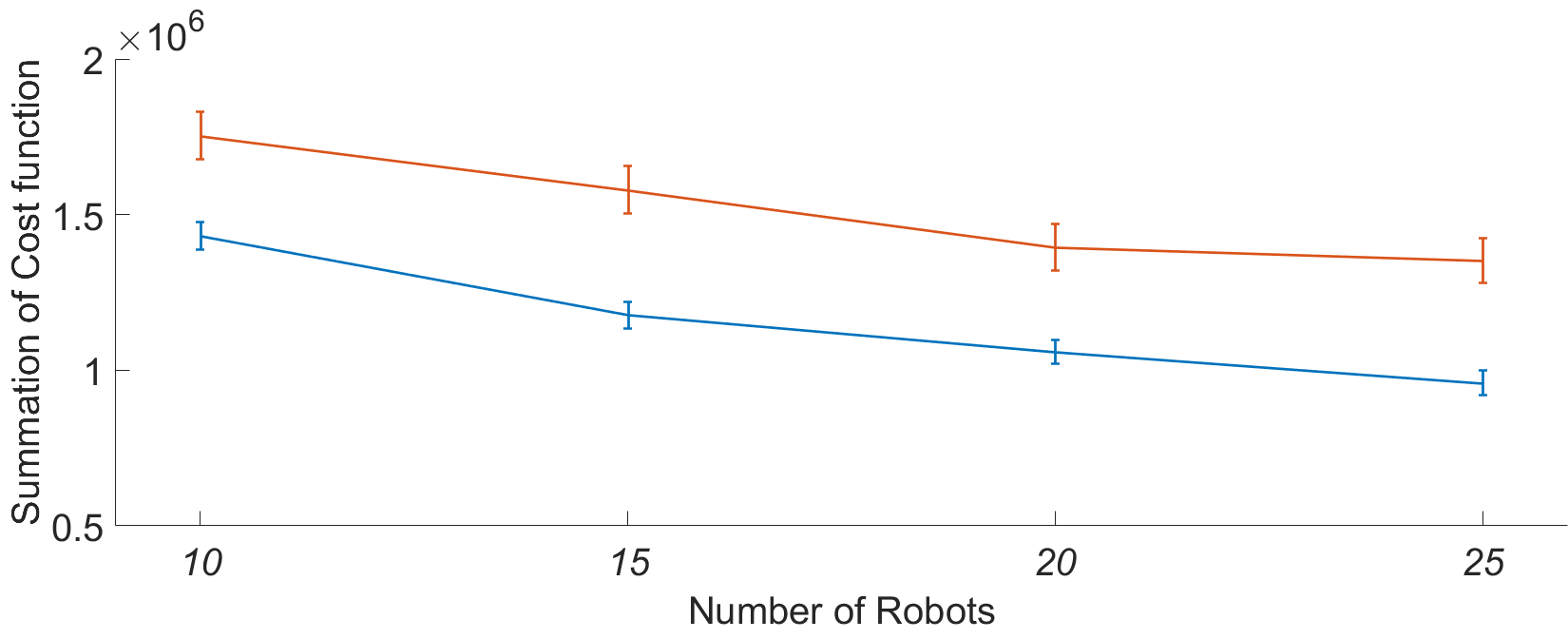}\label{fig:SumHalfx}}
		\caption{Comparison study for the \textit{right half-plane scenario}: proposed algorithm (blue) and approach presented in \cite{schwager2009decentralized} (red). (a) Final achieved value of the cost function. (b) Summation of the cost function over the experiment’s horizon.}
		\label{fig:SlotineRightHalfPlaneScenario}
	\end{figure*}
	
	As in the previous scenario, we present a comparison between the evaluated algorithms for different sizes of robot teams. The results are illustrated in Figure \ref{fig:SlotineRightHalfPlaneScenario}. Again, the proposed approach utilizes all the available team resources in order to achieve optimal robot configurations with small variance around the average values.

	\section{Three-dimensional surveillance of unknown areas}
	\label{sec:3Dcoverage}
	
	A more elaborate variation of the previously described set-up has been proposed by \cite{renzaglia2012multi} and applied in several domains (e.g., \cite{kapoutsis2015real, scaramuzza2014vision}).  Although the problem is again the optimal placement of robots in realtime, the details of the simulation set-up are important.  First and foremost, the robots are moving inside a 3D space (e.g., unmanned aerial vehicles). The terrain to be covered is considered an unknown, non-convex, 3D surface the formation of which may form an arbitrary number and shape of obstacles. Furthermore, a realistic model for the robots’ sensors is employed and utilized in all the simulation scenarios.
	
	\subsection{Problem definition}
	In this simulation testbed, the decision variables (\ref{eq:stataSpace}) represent the positions of the robots in 3D space, i.e., $\textbf{x} = \left[ x_1^\tau,\dots,x_N^\tau\right]^\tau$, where $x_i \in \mathbb{R}^3$. It is assumed that the area to be monitored is constrained within a rectangle in the $(\mbox{x, y})-$coordinates as 
	$$
	{\cal U} = \big\lbrace \mbox{x, y } | \mbox{ x} \in [\mbox{x}_{min},\mbox{x}_{max}], \mbox{y} \in [\mbox{y}_{min},\mbox{y}_{max}] \big\rbrace
	$$
	where $\mbox{x}_{min},\mbox{x}_{max},\mbox{y}_{min},\mbox{y}_{max}$ are real numbers that define the ``borders'' of the area of interest. Using the definition of ${\cal U}$, the area can be defined as a function that maps each point $(\mbox{x, y}) \in {\cal U}$ to a point $\mbox{z}=z(\mbox{x, y})$ (height of unknown terrain at $(\mbox{x, y})$). A point $q = (\mbox{x, y, z})$ of the terrain is \textit{visible} if there exist at least one robot so that:
	\begin{itemize}
		\item the robot $x_i$ and the point $q$ are connected by a line-of-sight;
		\item $\left\|x_i - q \right\| \leq \mbox{\textit{thres}} $, where \textit{thres} defines the maximum distance the $i$th robot can ``see''.
	\end{itemize}
	Given the robots configuration $\textbf{x}(k)$ at timestamp $k$, we let ${\cal V}$ to denote the \textit{visible} area of the terrain, i.e., ${\cal V}$ consists of all points $q \in {\cal U}$ that are \textit{visible} from the robots. 
	
	Furthermore, the measurements' model for all the robots admits the following form:
	\begin{equation}
		\label{eq:sensorsNoise}
		y_{x_i-q}=\left\{ \begin{array}{ll} \left\|x_i - q \right\| + h_\xi(x_i,q)\xi & \mbox{ if } q \in {\cal V} \\
			\mbox{undefined} & \mbox{ otherwise}
		\end{array}\right. \; \forall q \
	\end{equation}
	where $h_\xi(x_i,q)$ is the multiplicative sensor noise term $(e.g., \propto  \left\|x_i - q \right\|^2)$ and $\xi$ is a standard Gaussian noise. The above nonlinear noise model is a realistic representation of the noise effect in many real robot systems
	\cite[]{salavasidis2016terrain}\cite[Chapter 3-4]{teixeira2007terrain}. For instance, in the case of sonar or cameras, the noise affecting such sensors is proportional to the sensor-to-sensing-point distance, i.e., the larger the robot-to-sensing-point distance, the larger the sensor noise \cite[]{scaramuzza2014vision}.
	
	Having the above formulation in mind, we define the following combined cost function that the team of robots has to minimize
	
	\begin{equation}
		\label{eq:cf3dcover}
		\mbox{J}(\textbf{y}(k)) = \int_{q \in {\cal V}} \min_{i=1,\dots,N} y_{x_i-q} dq + K \int_{q \in {\cal U}\setminus{\cal V}}dq
	\end{equation} 
	
	The fist term is equivalent to the cost function considered in many coverage problems for known 2D environments \cite[]{cortes2002coverage, choset2001coverage}. The second term is related to the invisible area in the terrain. The positive constant $K$ serves as a weight for giving less or more priority to one of the objectives.
	
	Moreover, the set of nonlinear constraints (\ref{eq:constraints}), which must be held for each new robots' configuration $\textbf{x}(k)$,  include the following:
	\begin{itemize}
		\item the robots remain within the terrain's limits, i.e., within $[\mbox{x}_{min},\mbox{x}_{max}]$ and $[\mbox{y}_{min},\mbox{y}_{max}]$ in the x- and y-axes, respectively;
		\item the robots satisfy a maximum height requirement, while they do not hit the terrain, i.e., they remain within $[\mbox{z}+d_h,\mbox{z}_{max}]$ along the z axis, where $d_h$ denotes the minimum safety distance the robots should always have from the terrain and $\mbox{z}_{max}$ denotes the maximum allowable operational height for the robots;
		\item $\left\|x_i - x_j \right\| \geq d_r, \; \forall i,j \in \{1,\dots,N\} \mbox{ and } i \neq j $, i.e., the safety distance between two robots is $d_r$.
	\end{itemize}
	
	\begin{figure*}[!th]
		\centering
		\subfigure[]{\includegraphics[width=0.43\textwidth]{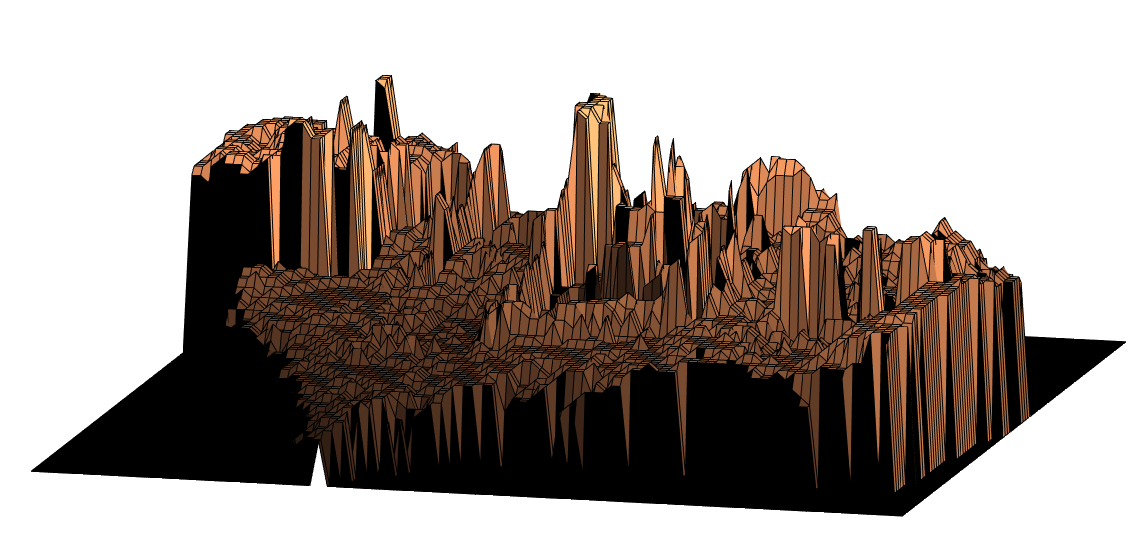}\label{fig:3d_coverage_1_3dEnviroment}}
		\subfigure[]{\includegraphics[width=0.43\textwidth]{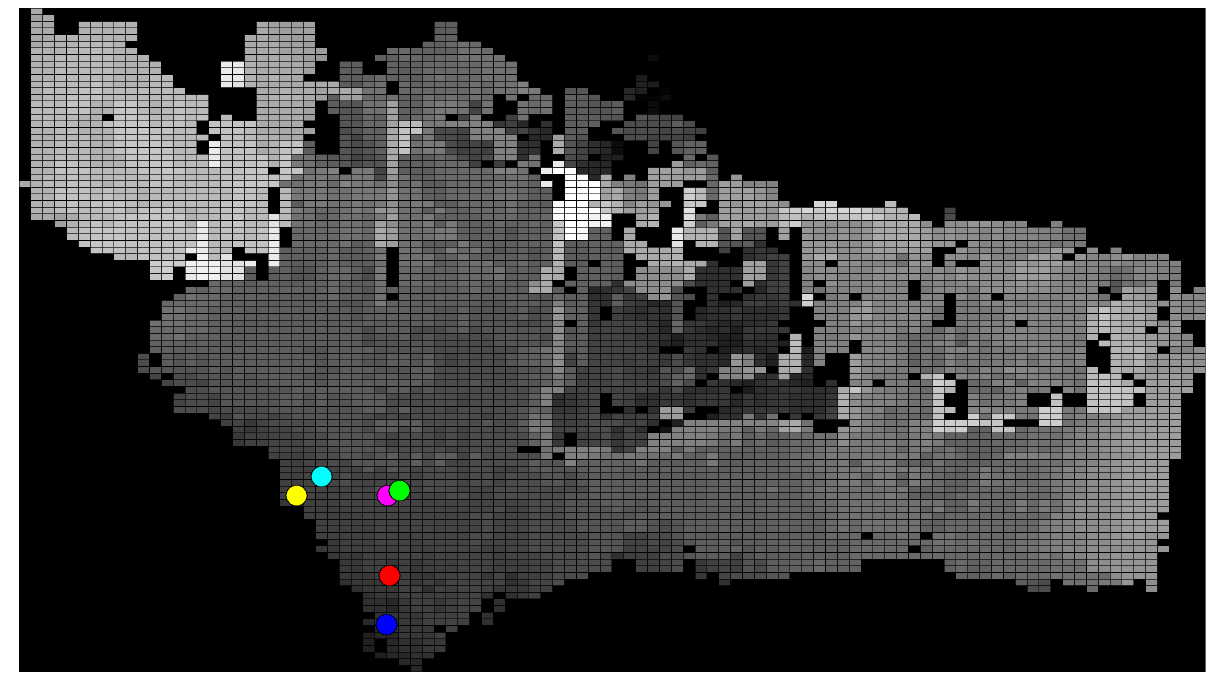}\label{fig:3d_coverage_1_3initPos}}	
		\subfigure[]{\includegraphics[width=0.43\textwidth]{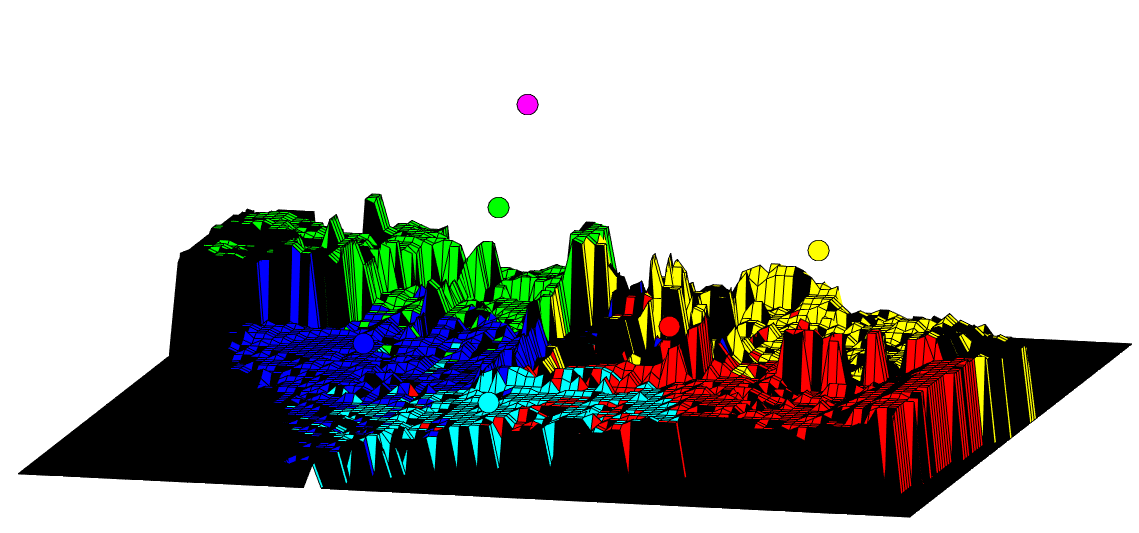}\label{fig:3d_coverage_1_CAO_2}}
		\subfigure[]{\includegraphics[width=0.43\textwidth]{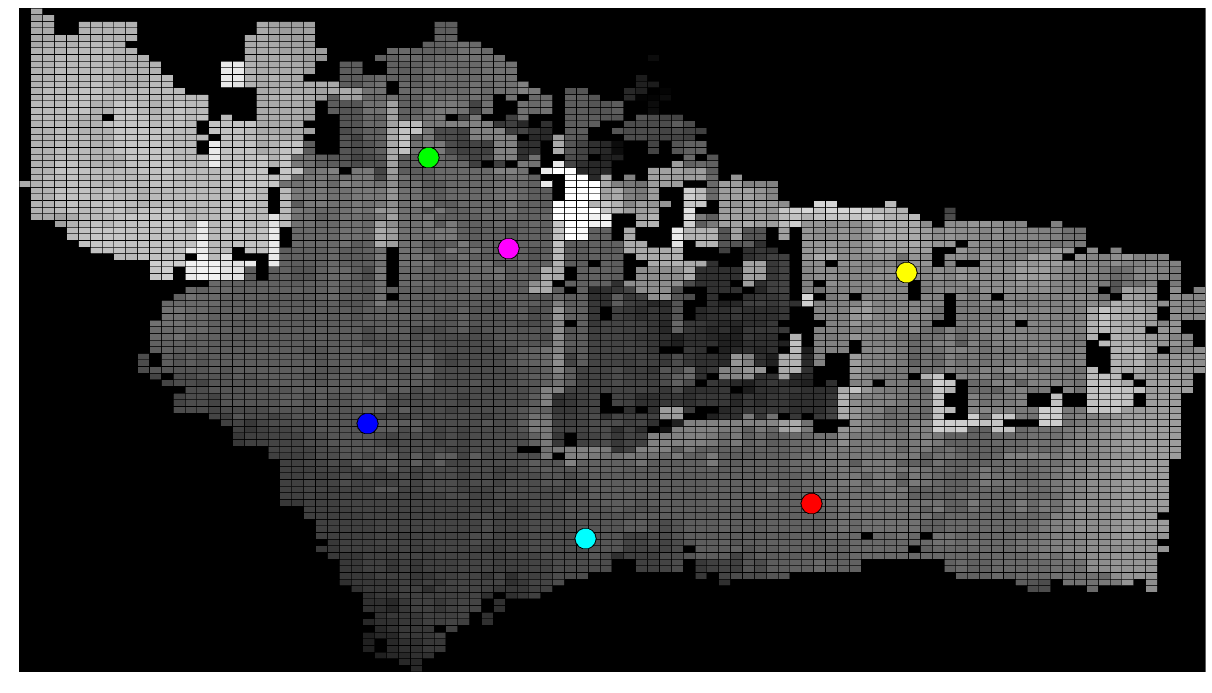}\label{fig:3d_coverage_1_CAO}}
		\subfigure[]{\includegraphics[width=0.43\textwidth]{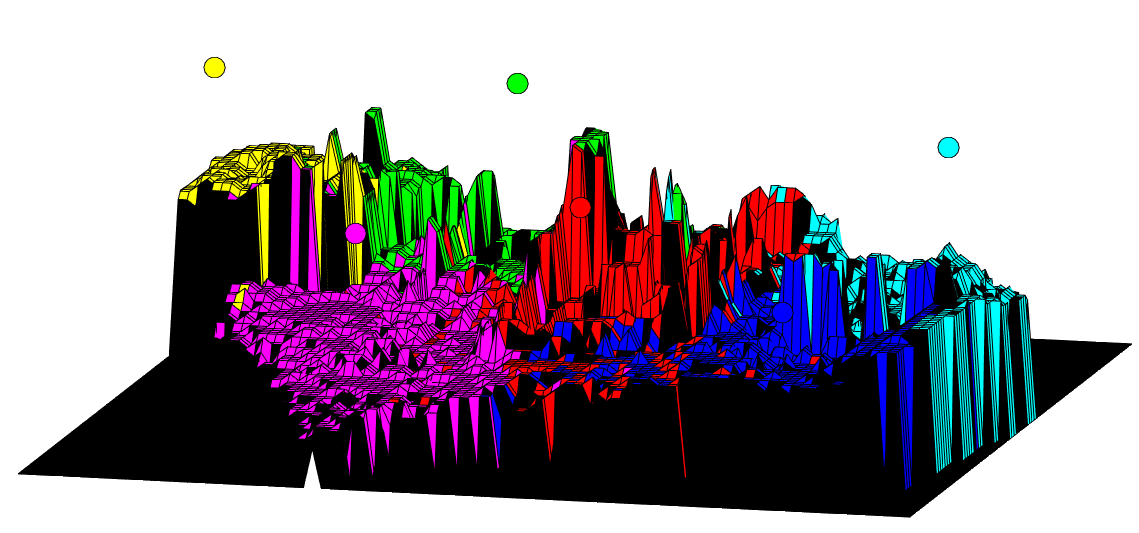}\label{fig:3d_coverage_1_Proposed_2}}
		\subfigure[]{\includegraphics[width=0.43\textwidth]{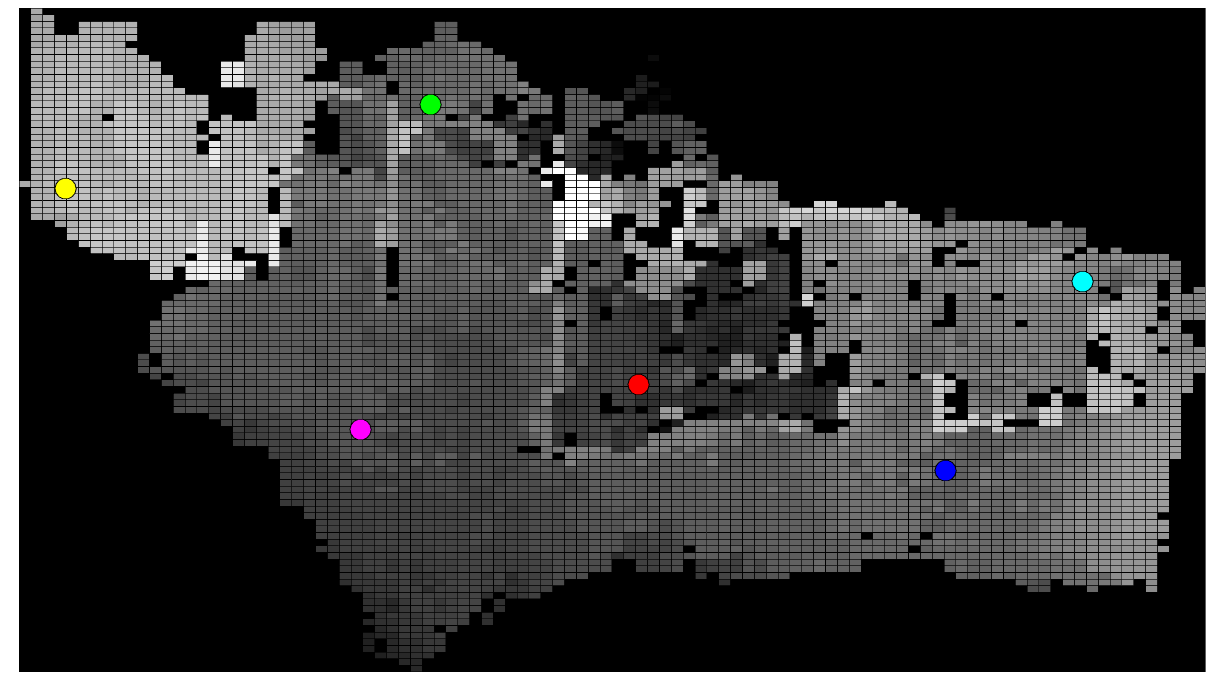}\label{fig:3d_coverage_1_Proposed}}
		\subfigure[]{\includegraphics[width=0.83\textwidth]{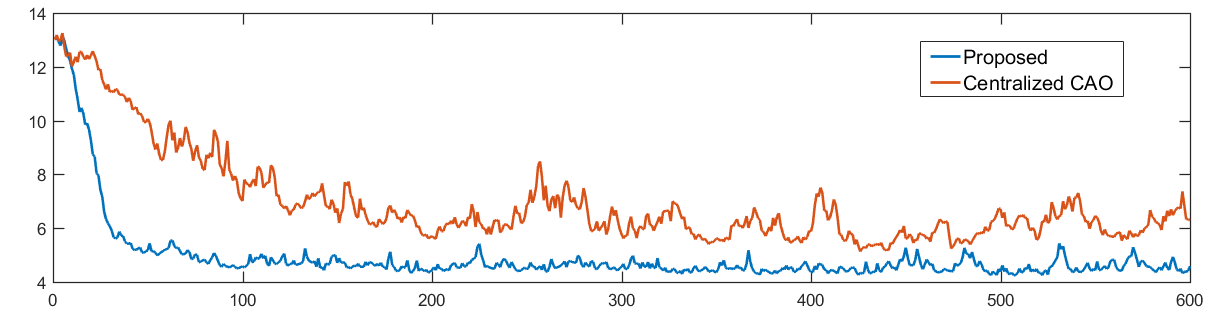}\label{fig:3d_coverage_1_3cf}}
		\caption{Surveillance of unknown terrain by a team of robots. The proposed algorithm and the CAO-based approach \cite[]{renzaglia2012multi} are evaluated on the same set-up (environment, robots initial positions, robots sensor capabilities). (a) 3D representation of the surface to be covered; (b) initial positions of the available robots; (c) 3D view, CAO-based approach; (d) top view, CAO-based approach; (e) 3D view, proposed approach; (f) top view, proposed approach; (g) cost function evolution.}
		\label{fig:3d_coverage_1}
	\end{figure*}
	\subsection{Simulation results} 
	The centralized CAO-based approach that has been proposed for the problem in hand \cite[]{renzaglia2012multi} is utilized for comparison purposes. The proposed approach was parametrized with a time window $T=40$ for the least-squares estimation, with $M=100$ random perturbations, the corresponding approximator was a third-order monomial estimator with $L = 18$, and the number of monomials per order (\ref{eq:monomials}) were $L_1 = 2, L_2=5 \text{ and } L_3 = 10$. Acknowledging the fact that the CAO algorithm performs optimization in a higher-dimensional space (centralized optimization scheme), a different set of parameters was chosen. Evaluating the CAO version for different numbers of random perturbations, we found that after $M=900$ the number of random perturbations does not affect its performance. Furthermore, to cope with the higher-dimensional state space, the time window was set to $T=60$ and the approximator was chosen to be a third-order monomial estimator with $L_1 = 3, L_2=12 \text{ and } L_3 = 40$ (with overall size of $L = 56$). In both algorithms, we utilize $\alpha=0.1$ to update the robot's positions. For the rest of this section, we use these values in all the presented experiments.
	
	To perform simulations in a realistic environment, we utilized the morphology of an area located in Z{\"u}rich, Switzerland (Figure \ref{fig:3d_coverage_1_3dEnviroment}). This map was generated using a state-of-the-art visual-SLAM algorithm \cite[]{doitsidis2012optimal}, which tracks the pose of the camera while, simultaneously and autonomously, building an incremental map of the surrounding environment. The terrain’s dimension is $[0, 162]$ m and $[0, 84]$ m for x and y axes, respectively, while the height of the terrain is between $[0, 7.2]$ m and the maximum operational height was set to $25$ m. Following the authors instructions \cite{renzaglia2012multi}, $K$ weight (\ref{eq:cf3dcover}) was chosen to be 30, whereas both the safety distance from the terrain and the minimum allowable distance between two robots were set to be $d_h=d_r=0.5m$. Finally, the duration of each experiment was set to $k_{\text{max}}=600$ timestamps. 
	
	Figure \ref{fig:3d_coverage_1} depicts such a simulation instance with six robots. The initial positions of the robots, as it is sketched in Figure \ref{fig:3d_coverage_1_3initPos}, were selected to be ``crowded'' inside a sub-area of the terrain. Figures \ref{fig:3d_coverage_1_CAO_2} and \ref{fig:3d_coverage_1_CAO} illustrate the final robots' configuration, as calculated by the CAO-based algorithm in 3D and 2D representation, respectively. The corresponding final robots' assignment as calculated by the proposed approach is presented in Figures \ref{fig:3d_coverage_1_Proposed_2} and \ref{fig:3d_coverage_1_Proposed}. In both cases, the 3D representation reports which sub-area of the terrain is covered by each \color{black} robot, whereas the 2D representation reveals the exact positions of the robots in $x-y$ plane and the distance between them. Figure \ref{fig:3d_coverage_1_3cf} depicts the evolution of the cost function (\ref{eq:cf3dcover}) for both the evaluated algorithms. Apart from the difference in the convergent state, the proposed approach is able to find this solution from its early steps $(<50)$. The centralized CAO needs more iterations to learn the dynamics of the robots and the unknown terrain, because it performs its optimization scheme in the higher-dimensional space of $\mathbb{R}^{3N}$ ($\mathbb{R}^{18}$ for the six robots).  In contrast, the proposed algorithm separately, although cooperatively, solves $N$(=6 robots for this instance) optimization problems of the size of $\mathbb{R}^{3}$. 
	
	In the specific problem set-up, the speed of convergence requires extra attention, as a slow convergence rate may lead to instability or loss of convergence at all. More specifically, if a navigation algorithm does not converge fast enough to the optimal configuration, one or more robots may have reached high-altitude positions, from which they cannot acquire useful measurements (out of their sensor capabilities (\ref{eq:sensorsNoise})). This is a non-recoverable situation, as the robots do not have any ``feedback'' from the terrain to properly evaluate their actions. 
	
	\subsubsection{Scalability analysis.}
	To validate both the efficiency and the effectiveness of the proposed algorithm in the case of larger robot teams, we performed experiments with 5, 10, 15, and 20 robots\endnote{Note that, for the current experiment set-up with the previously defined sensor's capabilities, the utilization of more than 15 robots cannot significantly affect the coverage task.}. For each different size of robotic team, we created 20 experiment instances with randomly chosen initial robots' positions. The aforementioned simulation instances are evaluated on both the proposed approach and the centralized CAO-based approach. 
	
\begin{figure*}[h]
		\centering
		\subfigure[]{\includegraphics[width=0.47\textwidth]{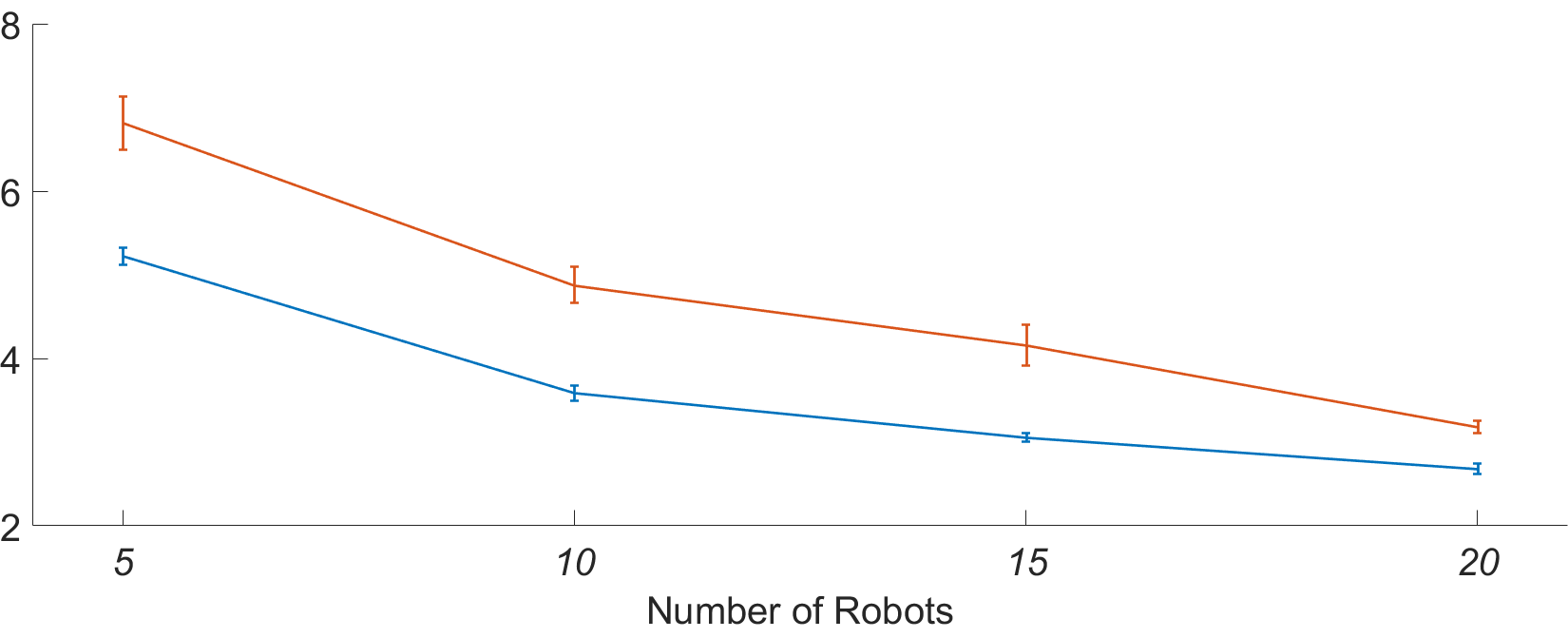}\label{fig:3d_scalability_cf}}
		\subfigure[]{\includegraphics[width=0.47\textwidth]{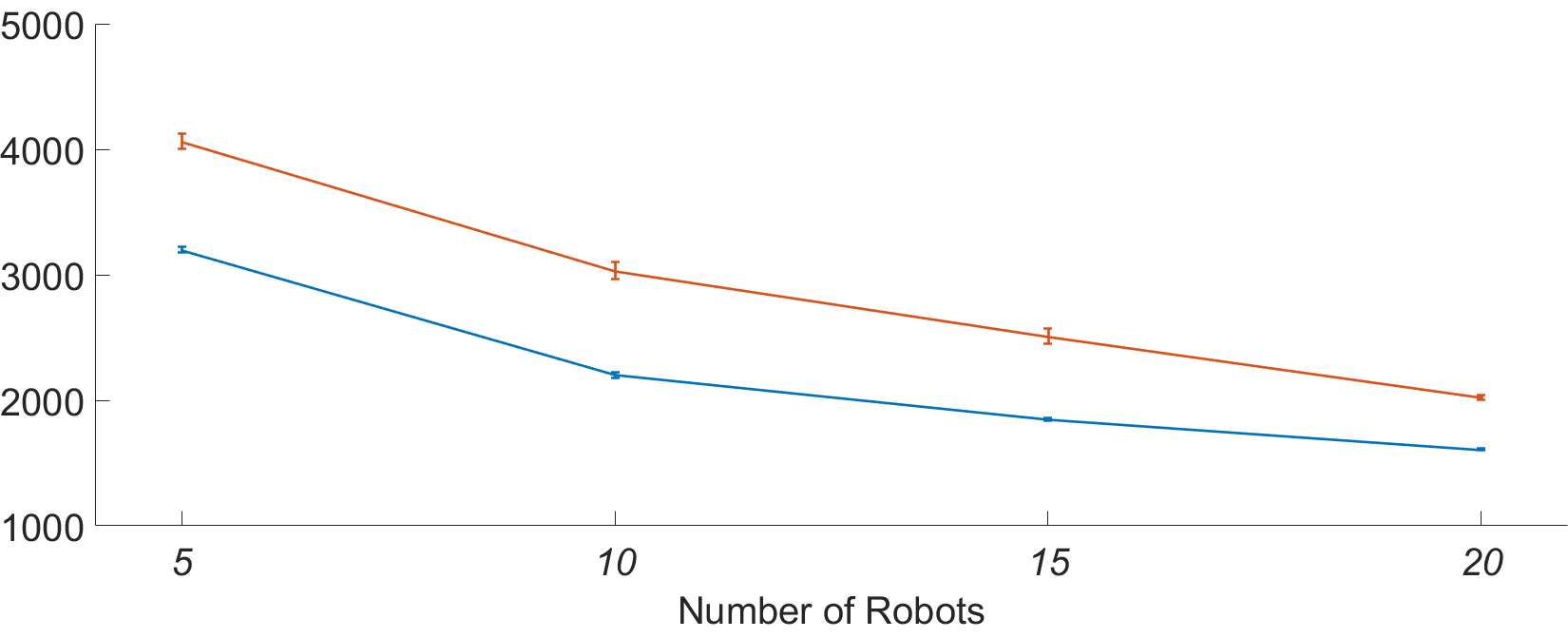}\label{fig:3d_scalability_SumCF}}
		\caption{Comparison study over different number of robots: proposed algorithm (blue) and CAO-based approach  \cite[]{renzaglia2012multi} (red). (a) Final achieved value of the cost function. (b) Summation of the cost function over the experiment’s horizon}
		\label{fig:3DscalabilityAnalysis}
	\end{figure*}

	The results of these simulations are summarized in Figure \ref{fig:3DscalabilityAnalysis}. Figure \ref{fig:3d_scalability_cf} displays the average value of the resulting cost function $\mbox{J}(k_{\text{max}})$, along with the corresponding confidence interval, over the different number of robots. In addition, Figure \ref{fig:3d_scalability_SumCF} displays a statistical analysis on the summation of cost function  $\sum_0^{k_{\text{max}}}{\mbox{J}}(k)$, to  investigate the convergence rate of each pair (scenario--algorithm). 
	
	Overall, the proposed approach achieves an average improvement of $23\%$ on the \textit{final achieved cost function value}, with $55.33\%$ improvement on the deviation around that average value. Moreover, the \textit{summation of cost function} has been improved by $23.84\%$ with a corresponding improvement on the deviation of $65.06\%$, against the centralized CAO-based approach. The proposed approach achieves these performance enhancements mainly due to the two following reasons.
	
	\begin{itemize}
	\item [(i)]  The proposed algorithm has a better perspective on the change of the overall cost function by evaluating the appropriate combinations of historical measurements on that cost function (\textit{Step 2-3} of the proposed approach).
	
	\item [(ii)] The fast convergence of the proposed approach eliminates the chances for a robot to be found out of its sensors capabilities. Therefore, the proposed approach is able to converge on approximately the same robots' configuration (per different team size), independently on the robots' initial positions. The latter is depicted in the substantial improvements on the corresponding confidence intervals.  
	\end{itemize}

	\subsubsection{Fault-tolerant characteristics.}
	In this scenario, we investigate the performance of the proposed algorithm in the case of catastrophic events or hardware failures. More precisely, five robots were initially deployed to perform the aforementioned coverage task, whereas the duration of the experiment was increased to $k_\text{max}=$1,000 timestamps. It is assumed that, at timestamp 330, one robot did not correspond to our control commands and the measurements' flow had been interrupted. Under these new circumstances, the surveillance task has to be undertaken by remaining, properly working robots. After the completion of two-thirds of the available timestamps, we assume that another robot had an equipment malfunction and cannot continue its covering task. Thus, the number of available robots, which are called to cover the area of interest for the $\sim$300 remaining timestamps, has dropped to three.
	
\begin{figure*}[!th]
		\centering
		\subfigure[]{\includegraphics[width=0.45\textwidth]{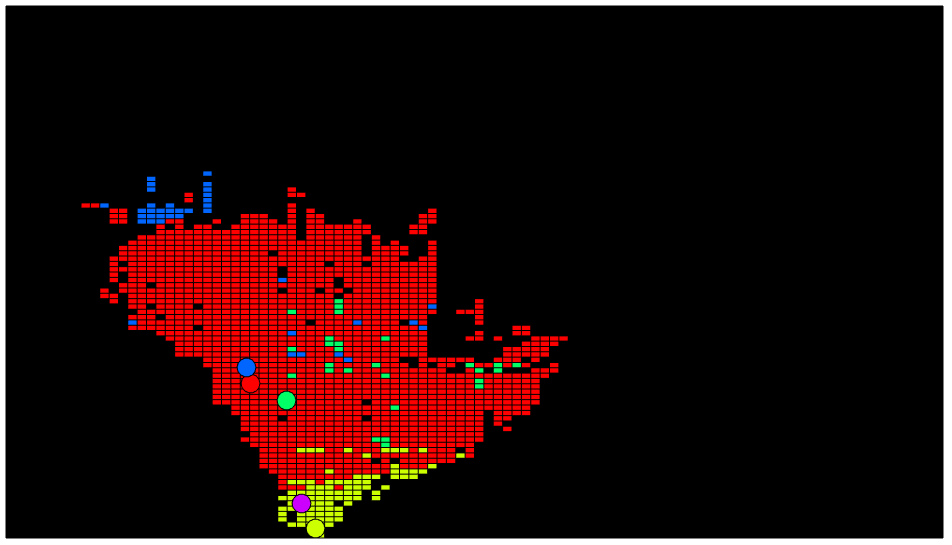}\label{fig:3d_coverage_F_3initPos}}
		\subfigure[]{\includegraphics[width=0.45\textwidth]{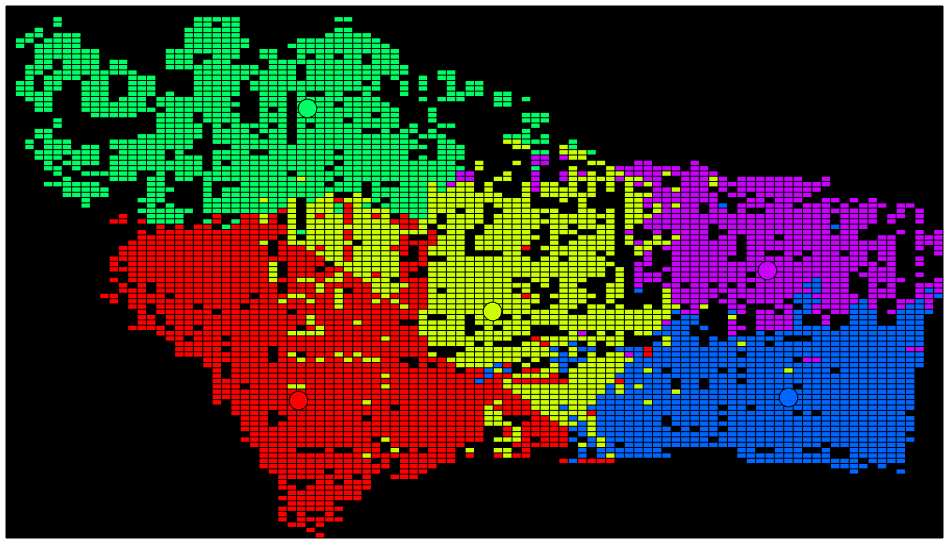}\label{fig:3d_coverage_F_330}}
		\subfigure[]{\includegraphics[width=0.45\textwidth]{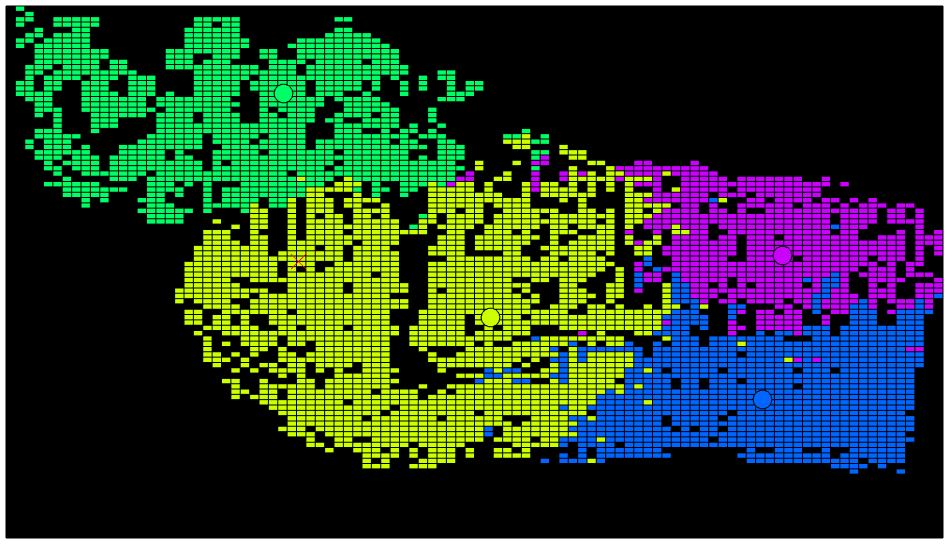}\label{fig:3d_coverage_F_331}}
		\subfigure[]{\includegraphics[width=0.45\textwidth]{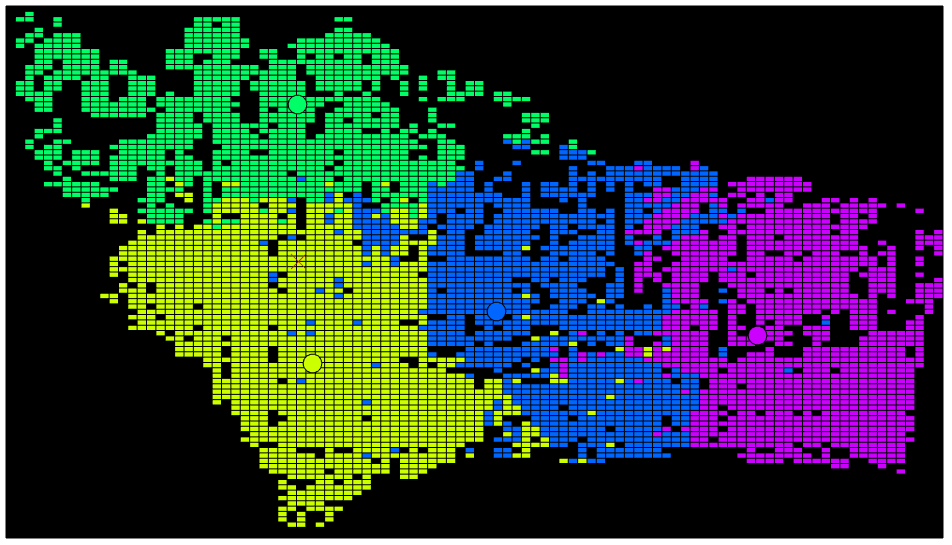}\label{fig:3d_coverage_F_660}}
		\subfigure[]{\includegraphics[width=0.45\textwidth]{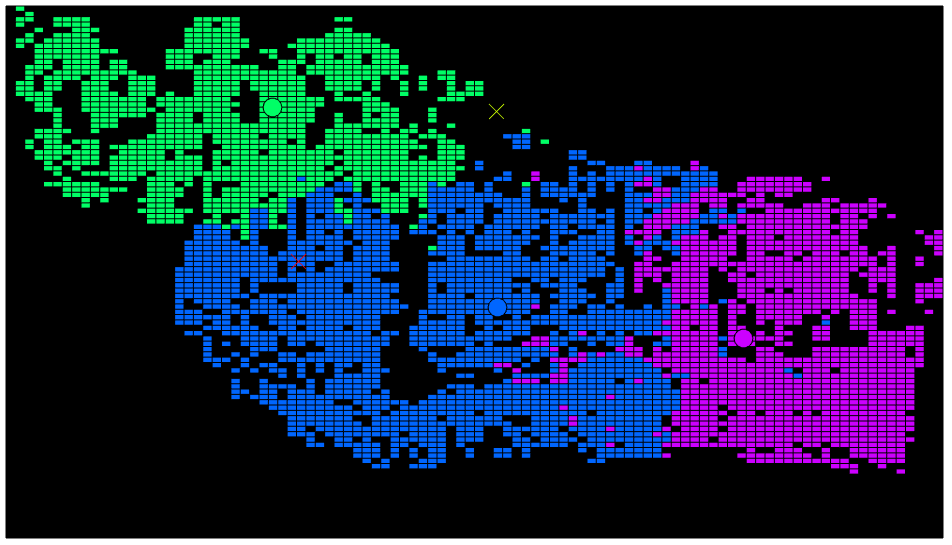}\label{fig:3d_coverage_F_661}}
		\subfigure[]{\includegraphics[width=0.45\textwidth]{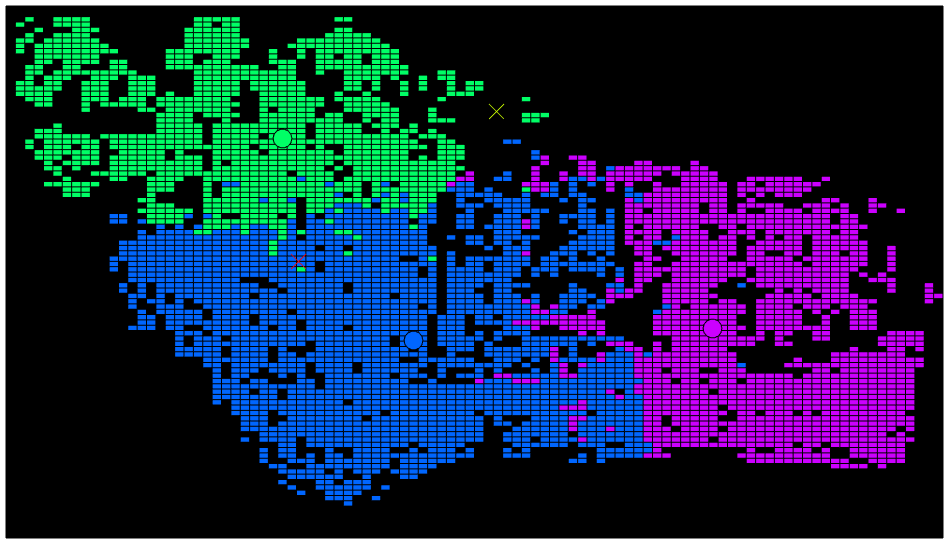}\label{fig:3d_coverage_F_Final}}
		\subfigure[]{\includegraphics[width=0.93\textwidth]{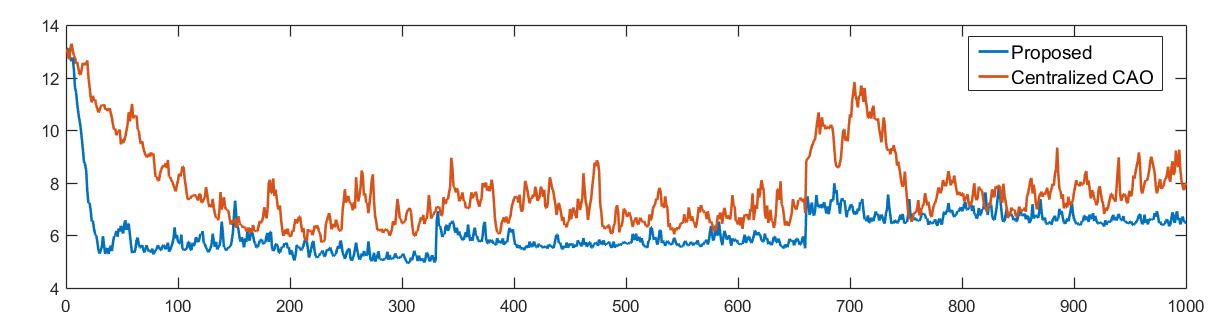}\label{fig:3d_coverage_F_3cf}}
		\caption{Malfunction scenario: five robots were initially deployed for the surveillance task. At two distinct timestamps, the swarm of robots loses one of its member due to a simulated malfunction. The surveillance task have to be continued with the remaining team resources. (a) Initial positions of the five available robots. (b) Coverage task with all five available robots. (c) One timestamp after the malfunction on the red robot. (d) Coverage task with four robots. (e) One timestamp after the malfunction on the yellow robot. (f)Again, the algorithm redesigns the robots positions to cover the area in the best possible way utilizing the available resources. (g) Cost function evolution.}
		\label{fig:3d_coverage_FaultTolerant}
	\end{figure*}

	Figures \ref{fig:3d_coverage_F_3initPos}--\ref{fig:3d_coverage_F_Final} illustrate the evolution of the robots positions during the course of the previously described scenario, utilizing the proposed approach. After both the robots' malfunctions, the algorithm redesigns the remaining robot positions to achieve the best possible coverage.  Overall, Figure \ref{fig:3d_coverage_F_3cf} demonstrates the evolution of the objective function for the proposed approach in comparison with the centralized CAO-based approach \cite[]{renzaglia2012multi}. 
	
	It must be emphasized that the proposed algorithm does not need any separately designed, fault-detection mechanism (e.g., failure in establishing communication, operator to detect the malfunction, etc.), as it is able to implicitly derive this kind of information from the changes in the cost function $\mbox{J}$ with respect to the commanded positions. The above feature is of paramount importance in real-life multi-robot applications, because it removes the tedious, and in many applications impossible, task to predict (or identify online) all the possible malfunctions, as well as to design the appropriate course of actions.

	\subsubsection{Target monitoring.}
	We close this section by investigating the algorithm's capability to process objectives that can be alternated/activated on the fly, without stopping and restarting the mission. To achieve this, simultaneously with the coverage task, we introduce the task of monitoring a target. For the sake of this simulation set-up, it is assumed that, in addition to the sensors which are responsible for the coverage task (\ref{eq:sensorsNoise}), the robots are equipped with exteroceptive sensors (e.g., cameras, sonars, etc.) which are able to estimate the targets' positions, according to the following measurement model:
	
	\begin{equation}
		\label{eq:measurmentOfATarget}
		y_{x_i-\chi_j}=\left\{ \begin{array}{ll} \left\|x_i - \chi_j \right\| + h_\xi(x_i,\chi_j)\xi & \mbox{ if } \chi_j \mbox{ \textit{has}} \\ & \mbox{ \textit{ been detected}} \\
			\mbox{undefined} & \mbox{ otherwise}
		\end{array}\right.
	\end{equation}
	
	where $\chi_j$ denotes the $j$th target's position in 3D space, $h_\xi(x_i,x_t)$ and $\xi$, similar to equation (\ref{eq:sensorsNoise}), denote the multiplicative sensor noise term and the standard Gaussian noise, respectively. Therefore, an extra term has to be added to the cost function (\ref{eq:cf3dcover}) to appropriately evaluate the progress of targets' monitoring, as follows:
	\begin{equation}
		\label{eq:cf3dcoverTT}
		\begin{aligned}
			\mbox{J}(\textbf{y}(k)) = \int_{q \in {\cal V}} \min_{i=1,\dots,N} y_{x_i-q} dq +  K \int_{q \in {\cal U}\setminus{\cal V}}dq \\ 
			+ K_t \sum_{j = 1}^{n_t}\min_{i=1,\dots,N} y_{x_i-\chi_j}
		\end{aligned}
	\end{equation}
	
	where $K_t$ serves as a weight to give more or less priority to the monitoring task in comparison with the coverage. In addition, $n_t$ denotes the number of targets to be monitored. 
	
\begin{figure*}[!th]
		\centering
		\subfigure[]{\includegraphics[width=0.95\textwidth]{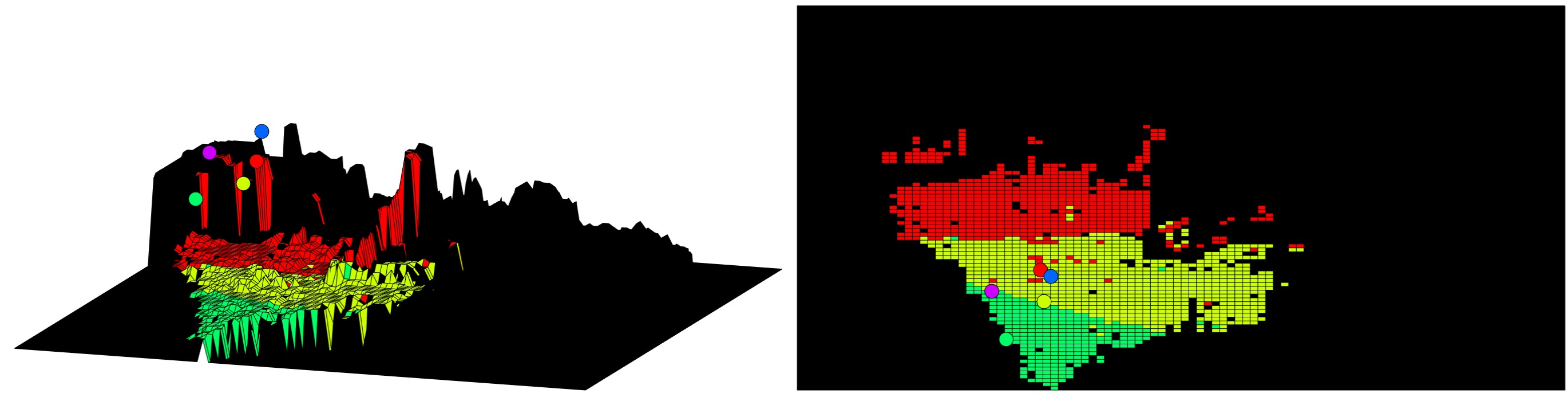}\label{fig:3d_coverage_2_3initPos}}
		\subfigure[]{\includegraphics[width=0.95\textwidth]{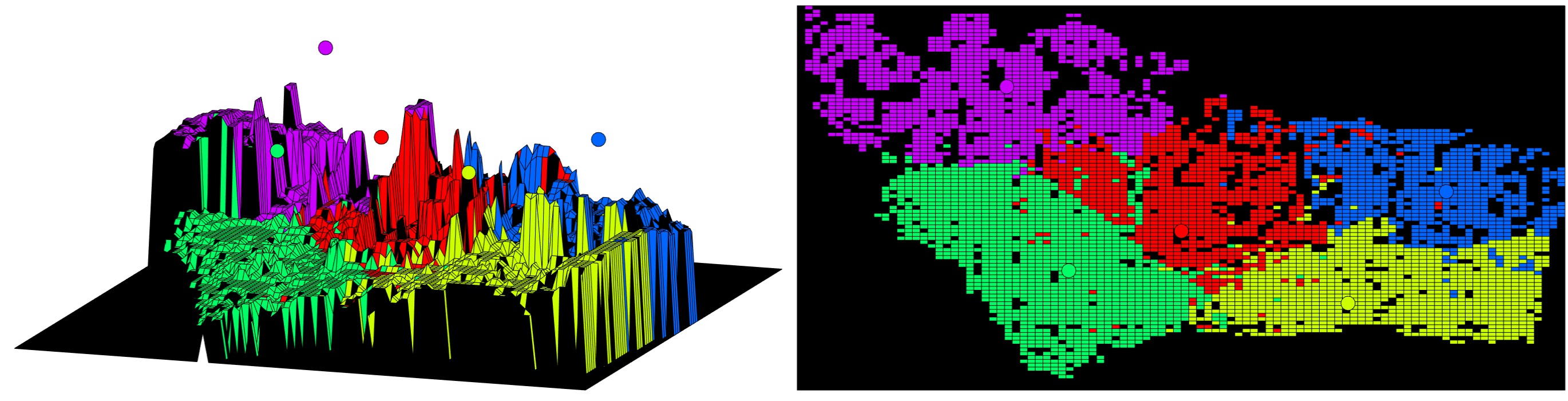}\label{fig:3d_coverage_2_300}}
		\subfigure[]{\includegraphics[width=0.95\textwidth]{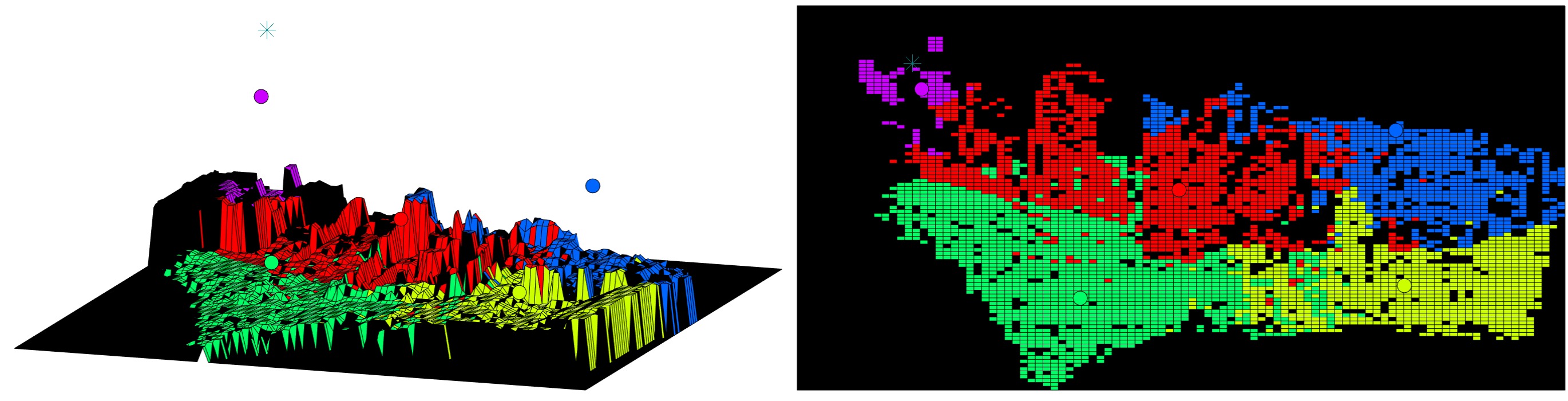}\label{fig:3d_coverage_2_target}}
		\subfigure[]{\includegraphics[width=0.95\textwidth]{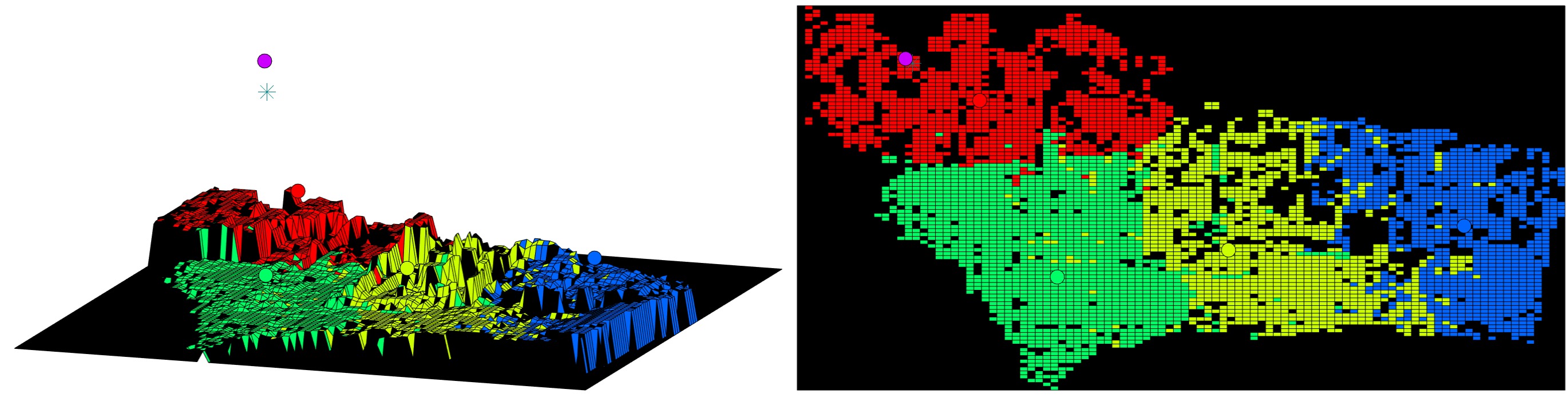}\label{fig:3d_coverage_2_Final}}
		\caption{Target monitoring scenario. The robots have been deployed with an extra objective (apart from the surveillance task) to get as close as possible to a target. The target appears inside the operation area of the robots in the middle of the mission. (a) Timestamp 1:initial positions of the five available robots. (b) Timestamp 367: coverage task with all five available robots. (c) Timestamp 427: the purple robot starts to gain height to minimize the distance from the target. As a consequence, it cannot cover adequately its underneath surface. (d) Timestamp 1,000: finally, the algorithm redesigns the robots positions so as to cover the area in the best possible way utilizing the available resources.}
		\label{fig:3d_coverage_TargetTracking}
	\end{figure*}

\begin{figure*}[]
		\centering
		\includegraphics[width=0.93\textwidth]{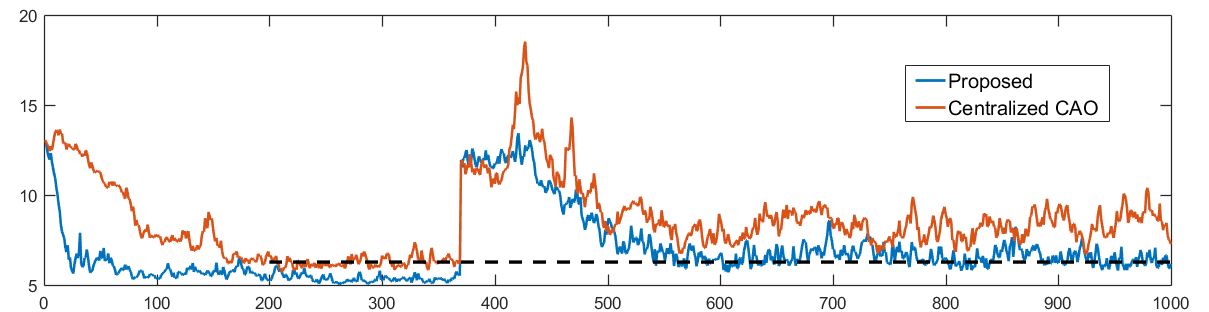}
		\caption{Cost function evolution in target monitoring scenario.}
		\label{fig:3d_coverage_2_3cf}
	\end{figure*}

	The experiments were performed in the same terrain, under the previously defined set-up parameters. Figure \ref{fig:3d_coverage_TargetTracking} illustrates four key snapshots, which demonstrate the functionality of the proposed algorithm. Figure \ref{fig:3d_coverage_2_3initPos} depicts the robots’ initial positions along with the corresponding coverage on the terrain. After 367 timestamps (figure \ref{fig:3d_coverage_2_300}), the algorithm has converged to the (locally) optimal robots' configuration for the coverage-only problem. At $k=370$ timestamp, it is assumed that a target, which requires closer examination, appears inside the operation area. The proposed algorithm, after the time needed to learn the changed problem dynamics (activation of the third term in (\ref{eq:cf3dcoverTT})), starts to adapt the robots' positions to minimize the updated cost function (\ref{eq:cf3dcoverTT}). More precisely, as illustrated in Figure \ref{fig:3d_coverage_2_target}, the purple robot (which was, at the time, closer to the target) starts to gain height to minimize its distance from the detected target. However, such an action leads to poor coverage on the subarea underneath that robot. To alleviate the above undesirable situation, the proposed algorithm redesigns the remaining robots' positions so as to achieve the best coverage of the terrain with the available resources. The final robots’ positions with the corresponding coverage of the terrain is sketched in Figure \ref{fig:3d_coverage_2_Final}. The evolution of the objective function for the proposed approach in comparison with the centralized CAO-based approach is demonstrated in Figure \ref{fig:3d_coverage_2_3cf}. Conclusively, for this simulation scenario, the proposed algorithm:
	\begin{itemize}
		\item chooses to assign a robot to be as close as possible to the target without any explicit command; 
		\item adapts the other robots positions so as to ``fill the hole'' in the coverage task; and
		\item achieves almost the same level of terrain coverage with the centralized CAO-based approach for five robots (Figure (\ref{fig:3d_coverage_2_3cf}) dashed line), whereas one (out of five) robots is occupied with another task.
	\end{itemize} 
	
	\section{Persistent coverage inside unknown environment}
	\label{sec:persistent}
	In the final application, we focus on the problem of persistent coverage in an area of interest with a team of robots. In this application, it is assumed that the operational robots are equipped with the appropriate sensors that are able to cover a portion of the environment.  The objective in a persistent coverage application is to continuously cover an area of interest, assuming that the coverage level follows a time-decaying function. \color{black} The problem along with a specifically designed algorithm has been proposed in \cite{palacios2016distributed}. The authors also established a well-defined, heuristic mechanism to online share the coverage evolution between the robots in a distributed way.
	
	Although the results are remarkable, the proposed decision-making mechanism in \cite{palacios2016distributed} utilizes a model that accurately predicts the improvement in the coverage level with respect to the robots movement \cite[equations (10),(18)-(21)]{palacios2016distributed}. In real-world applications, the above assumption does not always hold, as the increase in coverage level (i) is usually corrupted by nonlinear noise, (ii) can be affected by environmental specific characteristics, such as local morphology, obstacles, other robots' positions, etc., (iii) may follow a time-varying model (e.g., coverage level deteriorates over time). To circumvent these difficulties, we propose a variation of the above problem, where the changes in the level of coverage cannot be accurately predicted before the action. The actual information about the exact covered area is only available after the execution of each corresponding action through the robot's measurements. The above formulation is not only more \textit{realistic}, as it does not require an exact model of the environment or robot's coverage capabilities, but also more \textit{generic}, as it does not need to redesign the approach when robots with different or unknown coverage models are deployed. 
	
	\subsection{Problem definition}
	It is assumed that the operational area is a bounded $Q \subset \mathbb{R}^2$, which a team of robots has to persistently cover. The decision variables (\ref{eq:stataSpace}) represent the collective vector of all the robots' positions, i.e., $\textbf{x} = \left[ x_1^\tau,\dots,x_N^\tau\right]^\tau$,  where $x_i \in Q$. 
	
	Inside the environment there are several positions $q \in O \subset Q$ that cannot be traversed by the robots and additionally the presence of these obstacles affects each robot's coverage distribution. Although the exact positions of the obstacles are generally unknown, we assume that the robots are able to sense their presence when they are in close proximity. The above assumption is in line with the most commercial robots which are also equipped with proximity sensors to avoid collisions (e.g., \cite{nieuwenhuisen2014obstacle}). Thus, each robot's new candidate position $x_i^{\text{cand}}$ should verify the following constraint [see equation (\ref{eq:constraints}) of the general problem formulation]:
	\begin{equation}
		\label{eq:persistentConstraints}
		\min_{q \in O}\left( \left\| x_i^{\text{cand}} - q \right\| \right) \ge b
	\end{equation}
	where $b$ denotes the safety distance. At each timestamp $k$, the overall coverage increase is given by $y(q,k) = \sum_{i \in \{1,\dots,N\}} y_i(q,k),\; \forall q \in Q$, where 
	\begin{equation}
		\label{eq:persistentProduction}
		y_i(q,k) =\left\{ \begin{array}{ll}
			\gamma_i(q,x_i) & \mbox{ if } \left\|x_i -q \right\| \leq r_i^{\text{cov}} \text{ \& } \\
			& \mbox{there is line-of-sight} \\
			& \mbox{between } x_i \text{ and } q \\ 
			0 & \mbox{ otherwise}
		\end{array}\right.
	\end{equation}
	
	and $\gamma_i(q,x_i)$ denotes a nonlinear function that models how the coverage level evolves in the area around the $i$th robot's position. Note, that coverage distribution model $\gamma_i(q,x_i)$ may be different for each robot as it expresses the functionality of its on-board sensors.
	
	The coverage of the operational area can be modeled by a time-varying field and, in general, admits the following form:
	\begin{equation}
		\label{eq:persistentCoverageModel}
		Z(q,k) = d(q)Z(q,k-1)+y(q,k), \; \forall q \in Q
	\end{equation}
	
	In other words, the coverage level decreases to a constant decay gain $d(q)$, with $0<d(q)<1$, and increases according to the $y(q,k)$. The objective of the multi-robot team is to maintain a desired coverage level, $Z^*(q)>0,\; \forall q \in Q$. 
	
	Having the above formulation in mind, we define the \textit{quadratic coverage error} the robot team has to minimize
	\begin{equation}
		\label{eq:persistentCF}
		\mbox{J}(k) = \int_{Q} \left(Z^*(q) - Z(q,k) \right)^2  dq
	\end{equation} 
	
	\subsection{Simulation results} 
	
	All simulations were performed in a rectangle environment consisting of $100\times150$ units, with uniformly distributed decay rate $d(q)=0.995,\; \forall q \in Q$. The desired coverage level is $Z^*(q)=100,\; \forall q \in Q$. The number of robots was $N=6$, whereas their maximum motion is $u^{\text{max}}=5$. The coverage increase in \textit{open space} (obstacle-free), caused by the robots' movements, can be simulated by:
	\begin{equation}
		\label{eq:persistentModelCoverageIcrease}
		\gamma_i(q,k) = \frac{P}{r_i^{\text{cov}2}}\left(\left\|x_i - q\right\| - r_i^{\text{cov}} \right)^2
	\end{equation} 
	
	The maximum value is set to $P=17$ and the coverage radius is set to $r_i^{\text{cov}}=10$ units. Please note that this equation is not utilized during the decision-making process, but it is only employed to simulate the increase in the area coverage, with respect to the robot's movement. Finally, the experiments' duration is set to $k_{\text{max}}=900$ timestamps.
	
	To adapt the parameters of the proposed algorithm to the current application, we have to take into consideration that the navigation algorithm has to rapidly change its behavior owing to the time-varying nature of the cost function. Therefore, the time window for the least-squares estimation was only $T=5$ timestamps and the number of perturbations was $M=100$ candidates. To solve the underlying least-squares optimization problem (\ref{eq:estimatorTH}) with such a reduced historical values, we utilize only a second-order monomial estimator with $L_1 = 2 \text{ and } L_2 = 2$ (with overall size of $L = 5$).  Finally, following also the problem definition in \cite[Section II.]{palacios2016distributed}, we utilize $\alpha=1$ to update the robot's positions.
	
	\subsubsection{Obstacle-free environment.}
	\begin{figure*}[!th]
		\centering
		\subfigure[]{\includegraphics[width=0.32\textwidth]{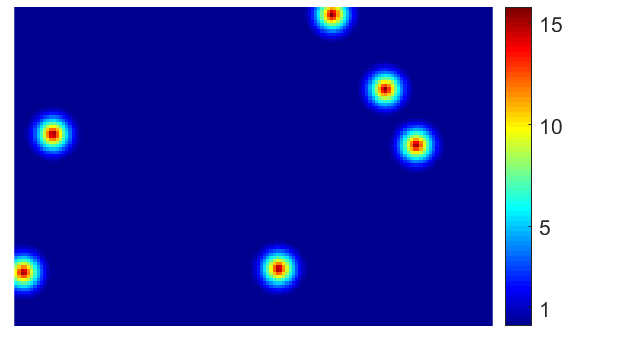}\label{fig:persistentCoverageEx1}}
		\subfigure[]{\includegraphics[width=0.32\textwidth]{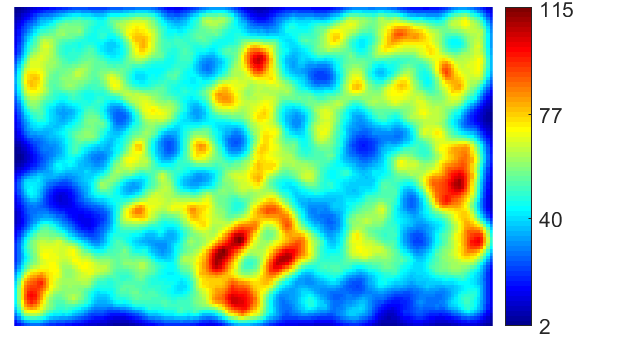}\label{fig:persistentCoverageEx2}}
		\subfigure[]{\includegraphics[width=0.32\textwidth]{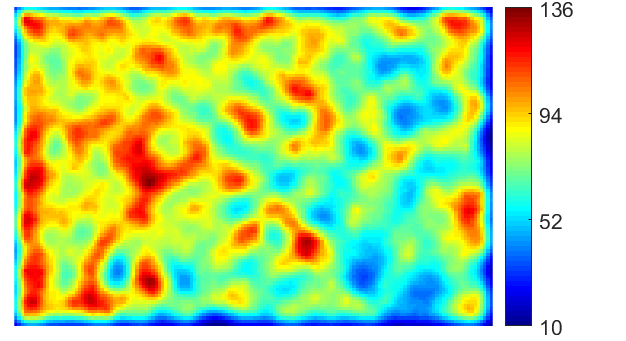}\label{fig:persistentCoverageEx3}}
		\subfigure[]{\includegraphics[width=0.32\textwidth]{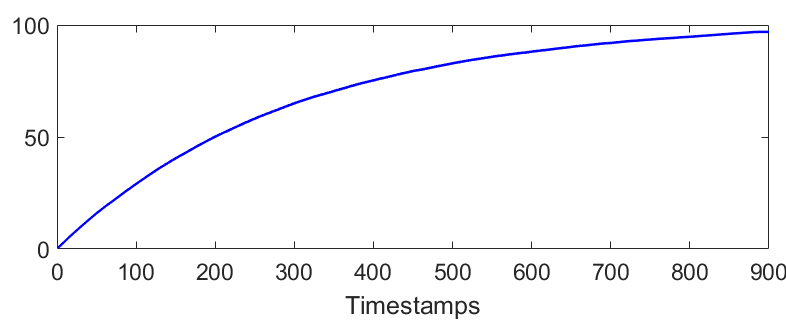}\label{fig:persistentCoverageExVals1}}
		\subfigure[]{\includegraphics[width=0.32\textwidth]{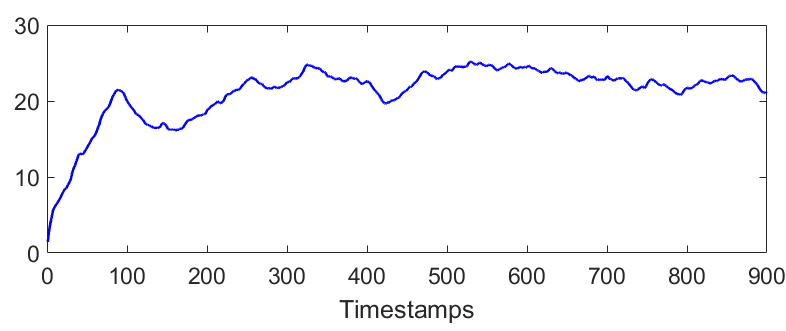}\label{fig:persistentCoverageExVals2}}
		\subfigure[]{\includegraphics[width=0.32\textwidth]{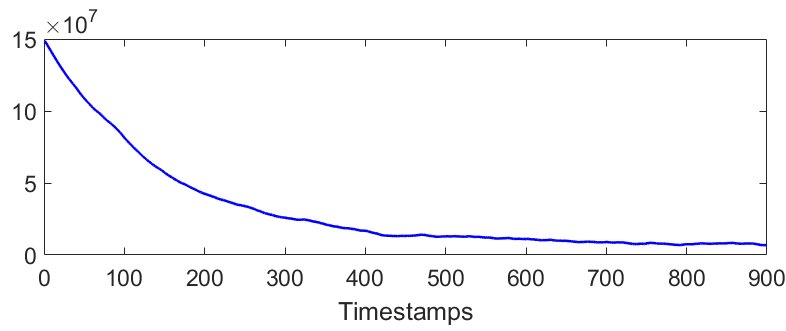}\label{fig:persistentCoverageExVals3}}
		\caption{Obstacle-free scenario: the coverage level for three different timestamps ((a) timestamp 1, (b) timestamp 200, and (c)timestamp 400) and the corresponding performance indices ((d) average coverage level, (e) standard deviation of coverage level, and(f) cost function, quadratic coverage error).}
		\label{fig:persistentCoverageExample}
	\end{figure*}
	
	In the first simulation scenario, we deploy the team of robots in an obstacle-free environment. An indicative simulation run of this scenario is summarized in Figure \ref{fig:persistentCoverageExample}. Figures \ref{fig:persistentCoverageEx1}--\ref{fig:persistentCoverageEx3} present the evolution of the coverage across the environment $Q$, for three different timestamps. In addition, Figures \ref{fig:persistentCoverageExVals1}, \ref{fig:persistentCoverageExVals2}, and \ref{fig:persistentCoverageExVals3} depict the evolution of the \textit{average coverage level}, the corresponding \textit{standard deviation}, and the \textit{quadratic coverage error} for the course of the experiment, respectively.  After the experiment execution, the average coverage level in all the operational environment $Q$ was $97$ with a standard deviation of $21.2$ and the corresponding quadratic coverage error was $6.9\times10^6$. 
	
	It should be highlighted that, the objective (\ref{eq:persistentCF}) is a time-varying function with high rate of change, i.e., the evaluation of (\ref{eq:persistentCF}) may result in significantly different scores for the same robots positions, even for very close timestamps. However, the proposed scheme is able to appropriately tackle the above problem, by constantly learning these cost function variations with respect to the robots' positions. 
	
	Although the proposed algorithm presents an equivalent performance compared with the dedicated one \cite[Section VI.]{palacios2016distributed}, if the problem is defined as in this scenario and the coverage evolution with respect to the robots movement being accurately predicted, a dedicated approach should be preferred to avoid the extra time due to learning (equations (\ref{eq:estimatorJ}) and (\ref{eq:estimatorTH}) of the proposed algorithm). However, the proposed approach has several advantages when it is deployed in a real-world environment, where the evaluation of the coverage increase cannot be performed beforehand. Such a scenario is presented in the following paragraph. 
	
	\subsubsection{Unknown cluttered environment.}
	
	\begin{figure*}[!th]
		\centering
		\subfigure[]{\includegraphics[width=0.32\textwidth]{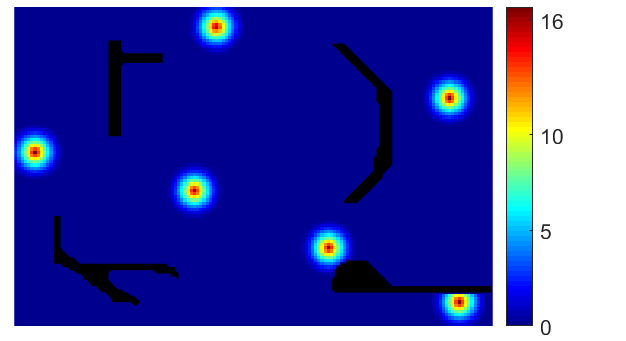}\label{fig:persistentCoverage2Ex1}}
		\subfigure[]{\includegraphics[width=0.32\textwidth]{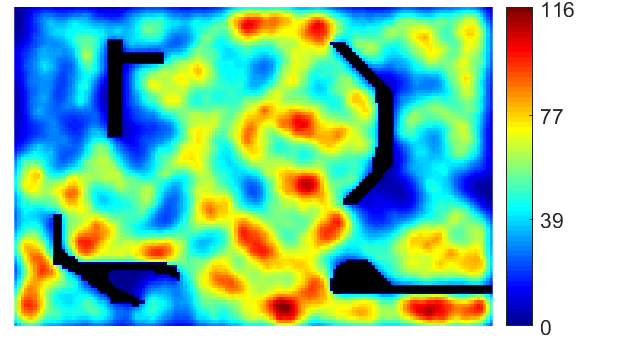}\label{fig:persistentCoverage2Ex2}}
		\subfigure[]{\includegraphics[width=0.32\textwidth]{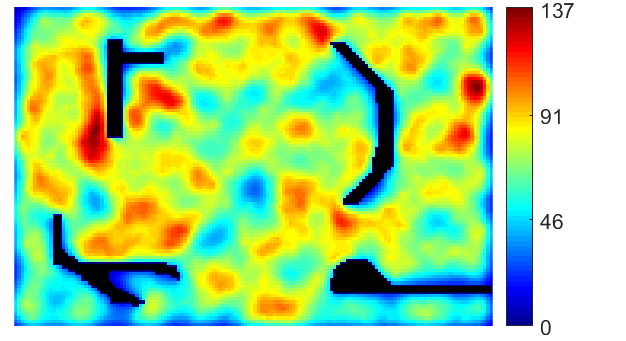}\label{fig:persistentCoverage2Ex3}}
		\subfigure[]{\includegraphics[width=0.32\textwidth]{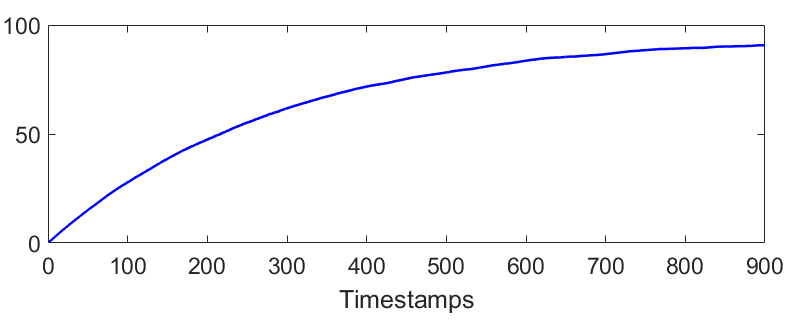}\label{fig:persistentCoverage2ExVals1}}
		\subfigure[]{\includegraphics[width=0.32\textwidth]{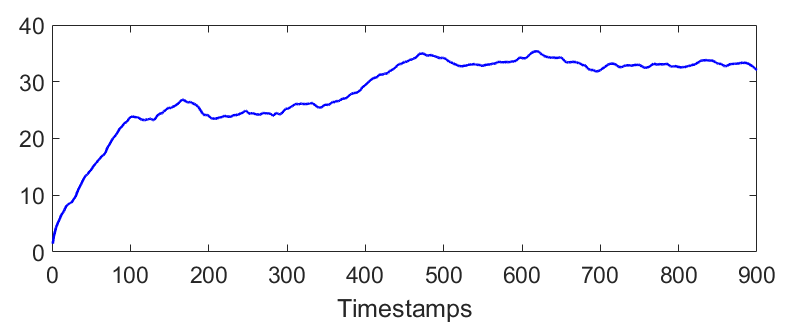}\label{fig:persistentCoverage2ExVals2}}
		\subfigure[]{\includegraphics[width=0.32\textwidth]{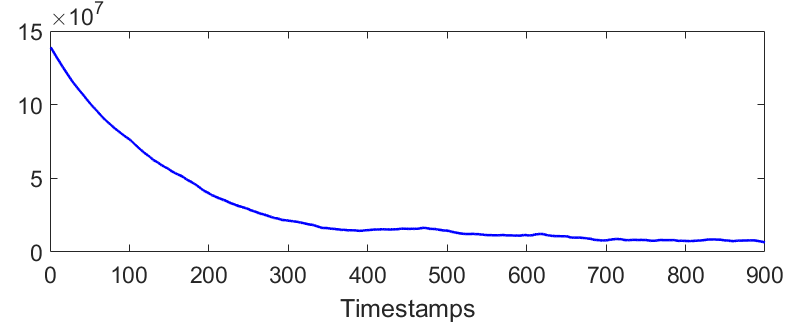}\label{fig:persistentCoverage2ExVals3}}
		\caption{Scenario in an unknown environment with non-convex obstacles: the coverage level for three different timestamps ((a)timestamp 1, (b) timestamp 200, and (c) timestamp 400) and the corresponding performance indices ((d) average coverage level, (e)standard deviation of coverage level, and (f) cost function, quadratic coverage error).}
		\label{fig:persistentCoverageExample2}
	\end{figure*}
	
	In the final simulation scenario, we investigate the performance of the proposed approach for the persistent coverage task, when it is evaluated on an unknown environment with non-convex obstacles. The obstacles have been created randomly and do not hold any kind of pattern. The minimum distance between the obstacles and any robot (\ref{eq:persistentConstraints}) has been set to $b=2.5$.
	
	Again, an illustrative example is presented in Figure \ref{fig:persistentCoverageExample2}. Following the same presentation policy, Figures \ref{fig:persistentCoverage2Ex1}, \ref{fig:persistentCoverage2Ex2}, and \ref{fig:persistentCoverage2Ex3} illustrate the evolution of the of the coverage across the environment $Q$, for three different timestamps. Figures \ref{fig:persistentCoverage2ExVals1}, \ref{fig:persistentCoverage2ExVals2}, and \ref{fig:persistentCoverage2ExVals3} depict the evolution of the \textit{average coverage level}, the corresponding \textit{standard deviation}, and the \textit{quadratic coverage error} (\ref{eq:persistentCF}), respectively.
	
	The cost function (\ref{eq:persistentCF}) does not need any adaptation to this scenario as the coverage values $Z(q)$ that correspond to obstructed locations $q \in O$ will remain zero, independently of their distance from any robot. In other words, the calculation of (\ref{eq:persistentCF}) does not need the information of the unknown obstacles, as the robots would never send coverage updates (\ref{eq:persistentProduction}) about the obstacles' positions. However, to construct comparable metrics with the previous scenario, we exclude the values that correspond to obstacles' locations from the calculation of the average coverage level (Figure \ref{fig:persistentCoverage2ExVals1}). After the experiment execution, the average coverage level inside $Q$ was $90.7$ with a standard deviation of $32.1$ and the corresponding quadratic coverage error was $6.6\times10^6$. 
	
	Comparing the outcomes of two scenarios side by side, we can draw the following observations:
	\begin{itemize}
		\item In the cluttered environment scenario, the robots can more easily get ``trapped'' in overcovered areas, resulting in a higher standard deviation. In other words, when a robot detects (implicitly from the changes in its corresponding cost function) that its position deteriorates the coverage level, may have only a small subset of possible new positions.
		\item During the course of the experiment in the cluttered environment, the obstacles ``blocked'' a portion of the robots' coverage capabilities. Therefore, for the cluttered environment scenario, the robots achieved a smaller average coverage level (excluding the obstacles positions).
	\end{itemize}
	
	\section{Conclusions}
	\label{sec:conclusions}
	A distributed methodology for dealing with multi-robot problems, where the mission objectives can be translated into an optimization of a cost function, has been proposed. In contrast to the majority of the multi-robot approaches, where the objectives are accomplished in a cost function optimization scheme, the proposed approach has been designed for multi-robot problems where the a priori calculation of the cost function is not feasible. In a nutshell, the proposed approach has the following key advantages:
	\begin{itemize}
		\item it does not require any knowledge of the dynamics of the overall system;
		\item it can incorporate any kind of operational constraint or physical limitation;
		\item it shares the same convergence characteristics as those of BCD algorithms;
		\item it has fault-tolerant characteristics; 
		\item it can appropriately tackle time-varying cost functions;
		\item and it can be realized in embedded systems with limited power resources.
	\end{itemize}
	
	Conclusively, we expect that many interesting tasks in mobile robotics can be approached by the proposed scheme. This is basically due to the fact that the proposed approach, instead of explicitly solving a particular problem, which requires prior knowledge of the system dynamics, learns, from the real-time measurements, exactly the features of the system which affect the user-defined objectives. Furthermore, the proposed approach can be appealing in many real-life application owing to its fault-tolerant characteristics, without an explicitly designed fault-detection mechanism. All the above issues are considered of paramount importance in the emerging field of multi-robot applications. 
	
	As future directions, we are interested in performing an extensive set of experiments, ideally with a large number of robots (e.g., a large swarm of femtosatellites (100 g class spacecraft) \cite{hadaegh2016development}). In particular, in such a set-up, it is impossible to explicitly program each and every robot to perform a subtask, therefore the goal will be to achieve an abstract set of objectives, which are defined in the form of cost function optimization. The idea behind the above formulation is, by excluding the intermediate steps from the design process, we enrich the multi-robot decision making scheme with autonomy, regarding the ``type'' of converged solutions. 
	

	\begin{funding}	
		This project has received funding from the European Research Council (ERC) under the European Union's Horizon 2020 research and innovation programme under grant agreement no 740593 (ROBORDER).
	\end{funding}

	\theendnotes
	
	\bibliographystyle{SageH}
	\bibliography{references}
	
\end{document}